\documentclass{article}
\PassOptionsToPackage{numbers,compress}{natbib}

  \usepackage[preprint]{neurips_2026}

\usepackage[utf8]{inputenc} 
\usepackage[T1]{fontenc}    
\usepackage{hyperref}       
\usepackage{url}            
\usepackage{booktabs}       
\usepackage{amsfonts}       
\usepackage{nicefrac}       
\usepackage{microtype}      
\usepackage{graphicx}
\usepackage[table]{xcolor}
\usepackage{amsmath}
\usepackage{amssymb}
\usepackage{tabularx}
\usepackage{array}
\usepackage{placeins}
\usepackage{listings}

\usepackage[most]{tcolorbox}

\usepackage{enumitem}

\usepackage{float}

\usepackage{algorithm}
\usepackage{algpseudocode}
\definecolor{mygreen}{RGB}{28,172,0}
\definecolor{purple}{RGB}{160,32,240}
\definecolor{navy}{RGB}{0,0,128}

\title{LLM-Driven Co-Evolutionary Automated Heuristic Design for Bi-Component Coupled Combinatorial Optimization}

\author{%
  Mingen Kuang\thanks{Equal contribution.} \\
  Xi'an Jiao Tong University \\
  \texttt{kuangme@stu.xjtu.edu.cn} \\
  \And
  Xudong Deng\footnotemark[1] \\
  Xi'an Jiao Tong University \\
  \texttt{dxd3125307059@stu.xjtu.edu.cn} \\
  \And
  Xi Lin \\
  Xi'an Jiao Tong University \\
  \texttt{xi.lin@xjtu.edu.cn} \\
  \AND
  Ye Fan \\
  Northwestern Polytechnical University \\
  \texttt{fanye@nwpu.edu.cn} \\
  \And
  Jianyong Sun \\
  Xi'an Jiao Tong University \\
  \texttt{jy.sun@xjtu.edu.cn} \\
  \And
  Jialong Shi\thanks{Corresponding author: \texttt{jialong.shi@xjtu.edu.cn}.} \\
  Xi'an Jiao Tong University \\
  \texttt{jialong.shi@xjtu.edu.cn} \\
}

\begin{document}

\maketitle

\begin{abstract}
While Large Language Models (LLMs) have recently shown promise in Automated Heuristic Design (AHD), existing methods typically generate and evolve heuristics as a single operator or search strategy, limiting their ability to model strong coupling among multiple decision substructures in problems such as the Traveling Thief Problem (TTP) and the Traveling Purchaser Problem (TPP). In this work, we propose CoEvo-AHD, an LLM-driven dual-population co-evolutionary framework for automated heuristic design in coupled combinatorial optimization. Unlike prior methods that evolve individual heuristics in isolation, CoEvo-AHD leverages LLMs to co-evolve two closely related operator populations. A cooperative evaluation mechanism explicitly captures interactions between route and selection operators, while pairwise scoring and synergistic joint crossover help discover complementary operator logic for joint improvement across coupled decision subspaces. We further design a tool-invocation environment library that encapsulates frequently used core operations, such as local-search delta computation, into callable functions, enabling LLM-generated operators to use standardized interfaces instead of reimplementing inefficient and error-prone problem-specific loops. Experiments on TTP and TPP show that CoEvo-AHD automatically discovers cooperative heuristic combinations and achieves competitive solution quality against traditional heuristics.
\end{abstract}

\section{Introduction}
\label{sec:introduction}

Large language models (LLMs) have recently opened a new direction for automated heuristic design, where executable heuristics can be generated, evaluated, and refined with limited human intervention. Existing LLM-driven methods have shown promising results in generating heuristic functions, construction rules, or local search operators through iterative loops of generation, evaluation, feedback, and regeneration \cite{romera2024funsearch,yang2024large,liu2024eoh,ye2024reevo}. However, most of these methods still treat a generated heuristic as an isolated algorithmic unit. This assumption becomes restrictive for many real-world combinatorial optimization (CO) problems, where a solution consists of multiple interdependent decision components and the effect of a local move cannot be assessed within a single component alone.

We use bi-component coupled CO to denote problems whose solutions can be decomposed into two semantically distinct components, while the feasibility or objective effect of modifying either component depends on the state of the other. In such problems, the utility of a component operator is inherently relational rather than intrinsic: it depends not only on how the operator modifies its own component, but also on how this modification interacts with the other component. Consequently, independently designing or evaluating operators for the two components may lead to moves that are locally admissible but misaligned with the joint objective.

The Traveling Thief Problem (TTP) \cite{bonyadi2013ttp,polyakovskiy2014benchmark} and the Traveling Purchaser Problem (TPP) \cite{manerba2017tppsurvey} are representative examples of this setting. They differ in how coupling is induced: TTP couples routing with load-dependent travel cost, where the tour order affects the cost of carrying selected items and the packing plan changes the evaluation of tour moves; TPP couples routing with market-dependent procurement feasibility and cost, where market selection, visiting order, prices, supplies, and demand satisfaction jointly determine solution quality. These problems illustrate a common challenge: high-quality search requires not only strong component-level operators, but also operator pairs whose behaviors are coordinated under the full coupled objective.

Although recent LLM-based studies have begun to explore heuristic sets and multi-operator collaboration \cite{liu2026eohs,zhao2026glns,qiu2026e2oc,kiet2026motif}, their focus is mainly on complementarity among algorithmic roles within general search frameworks, such as construction strategies, destroy--repair operators, or search-policy combinations. In contrast, bi-component coupled CO requires modeling complementarity among decision components of the solution itself. The key question is therefore not merely whether LLMs can generate stronger individual heuristics, but whether they can generate and refine paired component operators whose effectiveness is evaluated jointly under the full coupled objective.

To address this question, we propose \textbf{CoEvo-AHD}, an LLM-driven co-evolutionary automated heuristic design framework for bi-component coupled CO. CoEvo-AHD maintains two component-specific operator populations and uses LLMs to generate, rewrite, and recombine executable component operators. Instead of scoring operators in isolation, CoEvo-AHD embeds selected operator pairs into search trajectories over complete solutions, applies feasibility repair or component reconstruction when necessary, and evaluates the resulting solutions under the full coupled objective. The obtained rewards update both individual operator scores and pairwise synergy scores, aligning evolutionary selection pressure with the relational nature of operator utility. In addition to within-component mutation and crossover, CoEvo-AHD performs cross-component joint crossover on high-performing operator pairs, encouraging newly generated operators to share compatible communication protocols and complementary search assumptions.

CoEvo-AHD further introduces structured inter-component communication and a tool-augmented problem environment. The communication protocol specifies what types of information can be exchanged, such as search states, candidate modification suggestions, and local evaluation signals, without directly prescribing the heuristic logic for exploiting them. These signals are treated as verifiable search suggestions and can be accepted only after candidate construction, repair or reconstruction, and full-objective evaluation. The tool-augmented problem environment exposes reliable primitives for objective evaluation, feasibility repair, component reconstruction, marginal analysis, and delta evaluation, allowing LLM-generated code to focus on neighborhood design and cross-component search decisions rather than error-prone low-level implementation.

The contributions of this paper are threefold. First, we formulate LLM-driven automated heuristic design for bi-component coupled CO, where operator utility is relational and must be assessed through interactions between two heterogeneous solution components. Second, we propose CoEvo-AHD, a co-evolutionary framework with component-specific operator populations, paired execution-based evaluation, and pairwise synergy scores for selecting complementary operator pairs. Third, we design structured component communication, cross-component joint crossover, and a tool-augmented problem environment, and instantiate the framework on two representative problems, TTP and TPP, to evaluate its effectiveness.

\section{Related Work}
\label{sec:related_work}

\subsection{LLM-Driven Automated Heuristic Design}
\label{subsec:ahd}

Heuristic algorithms are widely used in combinatorial optimization, but their design often relies on manual specification of solution representations, neighborhood structures, and repair strategies. Recent advances in large language models (LLMs) have enabled LLM-driven automated heuristic design (LLM-AHD), where executable heuristic code can be generated and refined through program search. FunSearch combines LLMs with automatic evaluators to search for executable functions \cite{romera2024funsearch}. EoH formulates heuristic design as joint evolution of natural-language ideas and executable code \cite{liu2024eoh}, while ReEvo introduces reflective evolution that converts historical search performance into language feedback \cite{ye2024reevo}. HSEvo, MCTS-AHD, and MEoH further extend LLM-AHD in terms of population diversity, tree-search exploration, and multi-objective heuristic evolution \cite{dat2025hsevo,zheng2025mctsahd,yao2025meoh}.

These studies demonstrate that LLM-AHD can handle more complex algorithmic structures than single heuristics, but operator complementarity mainly occurs at the level of algorithmic roles. In contrast, our work binds operators to heterogeneous components of the complete solution, where their utility depends on interactions across components rather than the algorithmic workflow, thus modeling complementarity at the problem-structure level.

\subsection{Multi-Operator Collaboration and Positioning of This Work}
\label{subsec:multi_operator_ahd}

Recent studies have extended LLM-AHD to multi-operator collaboration. EoH-S generates complementary heuristic sets to improve cross-instance generalization \cite{liu2026eohs}. G-LNS co-evolves destroy and repair operators in a large neighborhood search framework and models their complementarity via cooperative evaluation \cite{zhao2026glns}. E2OC formulates interdependent operators in multi-objective evolutionary algorithms as sequential decision problems \cite{qiu2026e2oc}. MOTIF extends automated solver design from multi-strategy and multi-agent perspectives \cite{kiet2026motif}.

These studies demonstrate that LLM-AHD can handle more complex algorithmic structures than single heuristics, but operator complementarity is primarily confined to algorithmic roles within the search framework. In contrast, our work focuses on complementarity induced by the structure of the problem itself. Operators in CoEvo-AHD are tied to heterogeneous components of the solution, and their effectiveness depends on cross-component interactions rather than workflow roles. This positioning allows us to model problem-structure-induced complementarity through component-specific populations, paired execution-based evaluation, pairwise synergy scores, and cross-component joint crossover.

\subsection{Bi-Component Coupled CO: TTP and TPP}
\label{subsec:coupled_co}

Many real-world combinatorial optimization problems involve multiple interdependent subproblems. We use the Traveling Thief Problem (TTP) and the Traveling Purchaser Problem (TPP) as representative bi-component coupled CO benchmarks. TTP couples the Traveling Salesman Problem with the 0--1 Knapsack Problem: routing decisions determine the visiting order and items collected, while item selection affects travel speed and cost \cite{bonyadi2013ttp,polyakovskiy2014benchmark}. TPP jointly involves market selection, purchasing decisions, and route optimization to minimize the sum of travel and purchasing costs \cite{manerba2017tppsurvey}.

Existing algorithms for TTP and TPP include staged heuristics, memetic algorithms, ant colony optimization, tabu search, branch-and-cut, and ALNS variants \cite{faulkner2015approximate,mei2014matls,elyafrani2018population,wagner2016ants,wu2017exact,golden1981two,ong1982approximate,pearn1998improved,boctor2003heuristics,voss1996tabu,bontoux2008aco,goldbarg2009ta,laporte2003branchcut,yuan2024drl,kapancioglu2025ils,kucukoglu2024tpppf}. CoCo explicitly emphasizes coordination between routing and packing in TTP \cite{namazi2023coco}. However, most traditional methods rely on manually designed cross-component coordination mechanisms, which are problem-specific, hard to transfer, and require extensive tuning. In contrast, CoEvo-AHD leverages LLMs to automatically generate component-level operators and selects complementary operator pairs through paired execution-based evaluation and co-evolution, reducing dependence on handcrafted rules and expert intervention.

This positioning highlights that the novelty of CoEvo-AHD lies not only in leveraging LLMs for heuristic generation, but also in explicitly modeling and exploiting "problem-structure-induced operator complementarity" for bi-component coupled combinatorial optimization.

\section{Methodology}
\subsection{Overview of the Framework}
CoEvo-AHD consists of four stages: initialization, collaborative evaluation, population management, and LLM-driven evolution. In the initialization stage, separate operator populations are constructed for the two decision components. In the collaborative evaluation stage, pairs of component operators are jointly evaluated on training instances. The population management stage prunes low-performing operators, while the LLM-driven evolution stage generates new operators through intra-component mutation, intra-component homogeneous crossover, and cross-component collaborative joint crossover. Together, these stages form a closed loop of generation, evaluation, feedback, and regeneration, enabling the heuristic logic produced by the LLM to be continuously improved under real optimization feedback.

\begin{figure*}[t]
    \centering
    \includegraphics[width=0.95\textwidth]{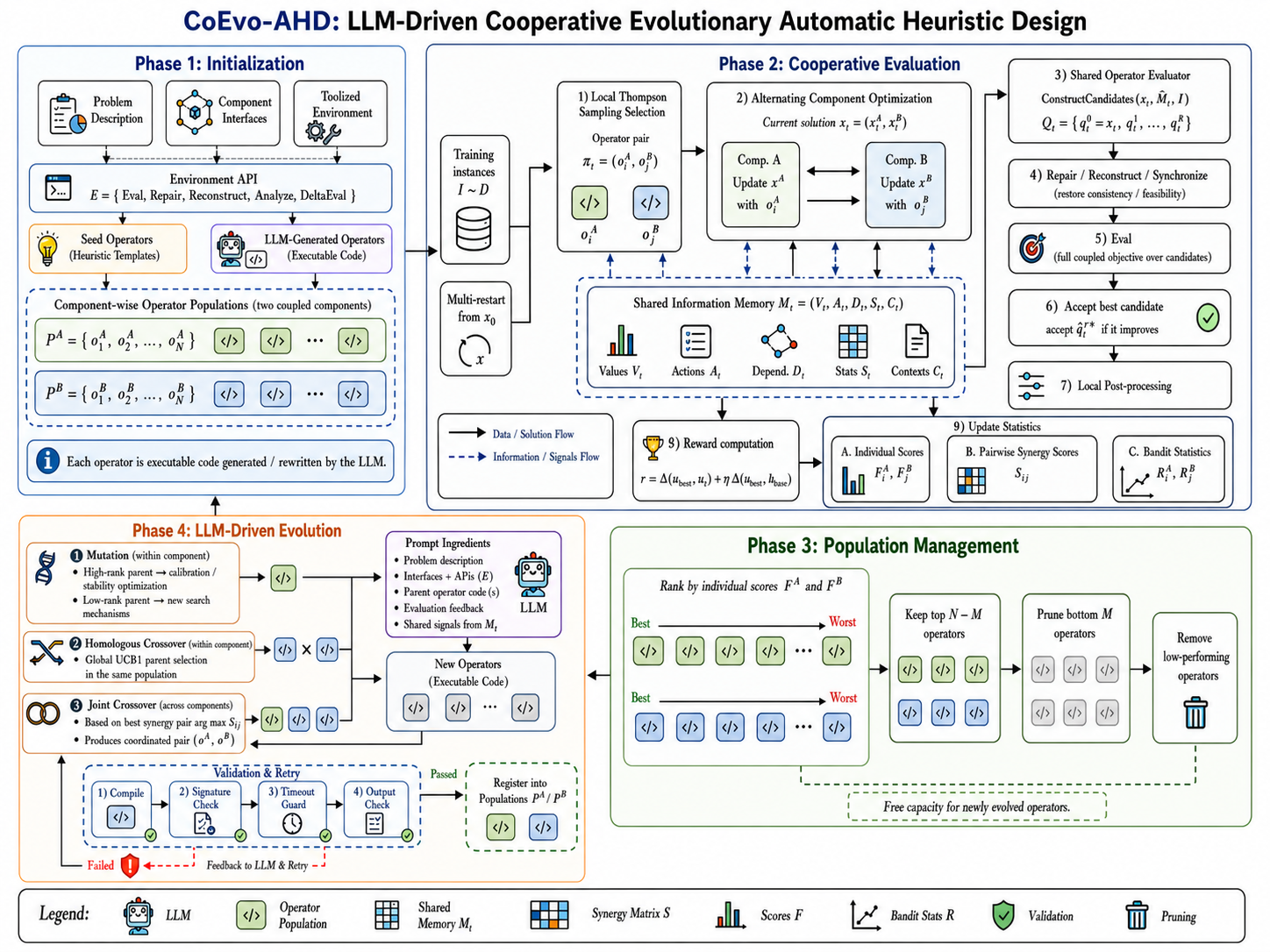}
    \caption{Overview of the CoEvo-AHD framework. The framework consists of four stages: initialization, collaborative evaluation, population management, and LLM-driven evolution. Two component-specific operator populations are maintained and co-evolved through pairwise evaluation, score update, pruning, mutation, homogeneous crossover, and cross-component joint crossover.}
    \label{fig:framework}
\end{figure*}

\subsection{Initialization}
\label{subsec:initialization}

For a bi-component coupled optimization problem, we denote a complete solution
as
\begin{equation}
    x=(x^{A},x^{B}),
\end{equation}
where $x^{A}$ and $x^{B}$ correspond to two heterogeneous but interdependent
decision components. CoEvo-AHD initializes two component-specific operator
populations:
\begin{equation}
    \mathcal{P}^{A}
    =
    \{o^{A}_{1},o^{A}_{2},\ldots,o^{A}_{N}\},
    \qquad
    \mathcal{P}^{B}
    =
    \{o^{B}_{1},o^{B}_{2},\ldots,o^{B}_{N}\}.
\end{equation}
For TTP, the two populations correspond to route operators and packing
operators. For TPP, they correspond to tour operators and purchasing operators.

The initial populations are constructed from two sources. First, we include a
small number of manually specified seed operators to provide valid executable
starting points and basic search behavior. Second, we use the LLM to generate
additional operators under the required component interface. Before entering
the population, every generated operator must pass syntax validation, interface
validation, bounded execution validation, and component-feasibility validation.
Invalid operators are discarded and regenerated.

In the implementation, each component operator returns only a candidate for its
own component:
\begin{equation}
    \widetilde{x}^{c}
    =
    o_i^c(x^{A},x^{B},\mathcal{I},\mathcal{M}),
    \qquad c\in\{A,B\}.
\end{equation}
The shared information space $\mathcal{M}$ is implemented as a mutable
dictionary, namely \texttt{problem\_data['shared\_info']}. Operators may update
this dictionary by side effect, and subsequent component operators or the
unified evaluator may read these shared fields when constructing candidate
complete solutions.

For each training instance, the framework also constructs an initial feasible
solution
\begin{equation}
    x_0=(x_0^{A},x_0^{B}).
\end{equation}
For TTP, $x_0^{A}$ is initialized by a TSP-style tour construction procedure,
and $x_0^{B}$ is initialized by a route-aware greedy packing procedure. For TPP,
$x_0^{A}$ is initialized as a feasible market tour, and $x_0^{B}$ is initialized
by reconstructing a purchasing plan over the visited markets. If a generated
candidate violates feasibility, the problem environment either repairs or
reconstructs the related component, or rejects the candidate if feasibility
cannot be restored.

The individual operator scores and pairwise collaboration scores are initialized
as
\begin{equation}
    F_i^{A}=0,
    \qquad
    F_j^{B}=0,
    \qquad
    S_{ij}=0,
    \quad
    \forall i,j.
\end{equation}
The local bandit statistics used for operator selection are also initialized
with empty reward histories. These statistics are updated only after operators
have been evaluated through complete-solution search trajectories.

\subsection{Cooperative Evaluation}
The cooperative evaluation stage estimates both the individual utility of each
component operator and the complementarity between operators from different
populations. For each training instance, CoEvo-AHD performs multiple
independent restarts. Each restart starts from a feasible initial solution and
iteratively applies one operator from population $\mathcal{P}^{A}$ and one
operator from population $\mathcal{P}^{B}$ within the same search trajectory.
Thus, operators are not evaluated in isolation, but through their contribution
to the complete coupled solution.

\paragraph{Local Operator Selection}
During instance-level evaluation, CoEvo-AHD uses a local Thompson Sampling
strategy to select one operator from each component population. Operators with
higher historical rewards are more likely to be selected, while newly generated
or rarely used operators still retain exploration opportunities. This produces
a component-operator pair $(o_i^A,o_j^B)$ for the current restart. The pair is
then evaluated jointly under the complete objective function, allowing the
framework to capture not only the quality of each operator but also their
cross-component compatibility.

\paragraph{Alternating Component Optimization}
Given a current solution $x_t=(x_t^A,x_t^B)$, the selected operators are invoked
alternately. Each operator receives the complete current solution, the problem
instance, and the shared information memory, but returns only a candidate update
for its own component. The framework then combines the updated component with
the unchanged counterpart to form a complete candidate solution. Since the two
components are coupled, the candidate is passed to the problem environment for
repair, reconstruction, validation, and complete-objective evaluation when
necessary. The candidate is accepted only if it improves the current complete
solution or restores feasibility in infeasible cases.

For TTP, this process alternates between route optimization and packing
optimization. A route operator proposes a new city permutation, which is
accepted only if the resulting TTP objective improves under the current packing
plan. A packing operator then proposes a new binary packing plan, which is
checked against the knapsack capacity and evaluated together with the current
route. After accepted improvements, a lightweight item-flip post-optimization
step is applied to further refine the packing component. For TPP, the same
procedure alternates between tour operators and purchasing operators, with the
environment ensuring demand satisfaction and purchasing feasibility.

\paragraph{Shared Information Memory}
To support cross-component coordination, CoEvo-AHD maintains a shared
information memory during each search trajectory. In our implementation, this
memory is realized as the mutable dictionary
\texttt{problem\_data['shared\_info']}. Component operators may write structural
signals into this memory, and subsequent operators may read them as soft search
guidance. For example, in TTP, route operators may record city positions, load
profiles, heavy-item cities, or segment-level priorities, which can then guide
packing operators. In TPP, tour operators may provide market insertion,
deletion, or replacement cues, while purchasing operators may expose demand gaps
or product-level marginal costs.

The shared memory does not directly modify the other component and does not
override the acceptance rule. Instead, it provides verifiable cues for candidate
generation. The final adoption of any shared suggestion is still determined by
feasibility checks and the complete coupled objective. This design preserves
the modularity of component-level operators while allowing search signals from
one component to influence the optimization direction of the other.

\paragraph{Tool-Based Problem Environment}
CoEvo-AHD further provides a tool-based problem environment to avoid requiring
LLM-generated operators to repeatedly implement low-level problem-specific
routines. The environment exposes stable and frequently used primitives, such
as objective evaluation, feasibility checking, repair or reconstruction, greedy
construction, structural analysis, and local move evaluation. In the TTP
implementation, for example, the environment provides callable interfaces such
as objective evaluation, fast 2-opt delta evaluation, and route-aware greedy
packing. These tools allow LLM-generated operators to focus on high-level
neighborhood design and component coordination, while relying on trusted
computational kernels for expensive or error-prone operations.

\paragraph{Reward and Statistics Update}
After each restart, CoEvo-AHD computes the reward according to the improvement
obtained by the selected operator pair over the initial solution of that
restart. The reward is used to update three types of statistics: the individual
score of the selected operator in $\mathcal{P}^{A}$, the individual score of
the selected operator in $\mathcal{P}^{B}$, and the pairwise synergy score of
the selected cross-component operator pair. The same reward is also recorded in
the local bandit statistics. These statistics are later used for local operator
selection, population pruning, and parent selection in the LLM-driven evolution
\subsection{Population Management}
After collaborative evaluation, the framework ranks the operator populations of the two components separately according to their individual operator scores. For each component $c\in\{A,B\}$, the top $N-M$ operators are retained:
\begin{equation}
    \mathcal{P}^{c}
    \leftarrow
    \mathrm{Top}_{N-M}
    (\mathcal{P}^{c};F^{c}),
    \quad c\in\{A,B\},
\end{equation}
while the lowest-ranked $M$ operators are pruned. The multi-armed bandit statistics associated with the pruned operators are removed accordingly. This process preserves high-performing operators while freeing population capacity for new operators generated by the LLM.
\subsection{LLM-Driven Evolution}
During the evolutionary stage, the LLM serves as a mutation and recombination operator in the code space. To refill the vacancies created by pruning, the framework employs three types of evolutionary operations: intra-component mutation, intra-component homogeneous crossover, and cross-component collaborative joint crossover.

\paragraph{Mutation}

Mutation generates a new operator from a single parent operator:
\begin{equation}
    \widetilde{o}^{c}
    =
    \Phi_{\mathrm{mut}}
    (o^{c}_{i},\mathrm{rank}(o^{c}_{i}),\mathcal{P}_{\mathrm{prompt}}),
    \quad c\in\{A,B\},
\end{equation}
where $\Phi_{\mathrm{mut}}$ denotes the LLM, and $\mathcal{P}_{\mathrm{prompt}}$ contains the problem description, component interface, available tools, parent code, and evaluation feedback. For high-ranking parents, mutation is biased toward parameter calibration and stability improvement; for low-ranking parents, mutation is encouraged to introduce new search mechanisms.

\paragraph{Homogeneous Crossover}

Homogeneous crossover is performed within the population of the same component. The framework selects two parent operators from the same component population and asks the LLM to recombine their search logic:
\begin{equation}
    \widetilde{o}^{c}
    =
    \Phi_{\mathrm{homo}}
    (o^{c}_{i},o^{c}_{j},\mathcal{P}_{\mathrm{prompt}}),
    \quad c\in\{A,B\}.
\end{equation}
Parent selection follows a global UCB1 rule. For a candidate parent $o_i$, its UCB1 value is defined as
\begin{equation}
    \mathrm{UCB1}(o_i)
    =
    \bar{r}_i
    +
    c_0
    \sqrt{
    \frac{\log N_{\mathrm{tot}}}{n_i}
    },
\end{equation}
where $\bar{r}_i$ is the average reward, $n_i$ is the number of selections, $N_{\mathrm{tot}}$ is the total number of selections over the candidate set, and $c_0$ is the exploration coefficient. Operators that have not been selected before are prioritized for exploration.

\paragraph{Collaborative Joint Crossover}

Collaborative joint crossover selects the best-performing pair of component operators according to the paired collaboration score:
\begin{equation}
    (i^{\star},j^{\star})
    =
    \arg\max_{i,j}
    S_{ij}.
\end{equation}
The LLM then rewrites this operator pair as a unified entity:
\begin{equation}
    (\widetilde{o}^{A},\widetilde{o}^{B})
    =
    \Phi_{\mathrm{joint}}
    (o^{A}_{i^{\star}},o^{B}_{j^{\star}},
    \mathcal{P}_{\mathrm{prompt}}).
\end{equation}
This mechanism encourages the newly generated component operators to share a consistent communication protocol, search assumptions, and complementary logic, thereby improving the coordination capability between paired components.

\section{Experiments}
To validate the effectiveness of the proposed CoEvo-AHD framework on coupled combinatorial optimization problems, we evaluate it on TTP and TPP, both of which involve two interdependent decision-making subproblems: route planning and resource selection. In TTP, the visiting route determines the time cost of carrying selected items, while the packing decisions affect the travel speed through the knapsack weight. Similarly, in TPP, procurement decisions and the visiting order jointly determine the overall cost.

During the evolutionary phase, we use randomly generated instances. At test time, the learned operators are evaluated on a separate set of randomly generated instances to assess their generalization ability. To mitigate the effect of stochasticity, we conduct three independent evolutionary runs for each task, each for 200 generations. The computational budget during evolution is limited to 100 iterations or 60 seconds of runtime. In the final testing phase, the best evolved pair of operators is applied to test instances under a fixed budget of 500 iterations or 100 seconds, and the resulting solution quality is used to evaluate the effectiveness of the learned operators.

\paragraph{Setting}
For the large language model used to generate and evolve operators, we adopt Qwen3.5-Plus. For the two co-evolving populations, we maintain a population size of N=5. During population management, the two individuals with the lowest fitness in each population are removed. In the evaluation process, each instance is run with three restarts by default.

For the TTP experiments, the training instances are automatically generated by an instance generator. By default, we use 16 training instances, each consisting of 50 cities, where each non-depot city is associated with one item. The knapsack data type is set to bounded-strongly-corr. CoEvo-AHD maintains two operator populations, corresponding to the route optimization operator and the pack optimization operator.

For the TPP experiments, the training and test instances are automatically generated in the TPPLIB-compatible Class 3 format \cite{RieraLedesma2012TPPLIB}, following the Euclidean instance generation scheme introduced by Laporte et al. \cite{laporte2003branchcut}. By default, we generate 10 instances for each of 11 city-product scales, ranging from $20 \times 20$ to $150 \times 150$, and split them into training and test sets with a ratio of 3:7. Each instance has unit product demand and unit supply quantity, with product availability probability set to 0.8 and prices sampled from $[1,100]$. CoEvo-AHD maintains two operator populations, corresponding to the tour optimization operator and the purchasing optimization operator. All the initial populations are constructed from both manually designed seed operators and LLM-generated operators.
\paragraph{Baseline}
To evaluate the performance of CoEvo-AHD, we compare it against several classical baselines. For TTP, we select three top-performing algorithms from the literature: MATLS\cite{mei2014matls}, S5\cite{el2018efficiently}, and CoCo\cite{namazi2023coco}. For TPP, we use DMD-ATA (ACO)\cite{bontoux2008aco}, TA\cite{goldbarg2009ta}, and ALNS\cite{kucukoglu2024tpppf} as the three baseline methods.
\subsection{Experimental Results}

Table \ref{TTPResult-table} reports the average objective values achieved by different methods on the TTP benchmark instances, where higher values indicate better solution quality. CoEvo-AHD exhibits strong performance on small- and medium-scale instances: it achieves the best results on TTP20 and TTP50, with objective values of 583.10 and 1730.60, respectively, outperforming all baselines. This suggests that the dual-population co-evolutionary mechanism can effectively capture the coupling between route planning and packing decisions, enabling the discovery of high-quality cooperative operator pairs in relatively small search spaces.

On larger instances, namely TTP100 and TTP200, CoEvo-AHD obtains the second-best results, trailing only CoCo. In terms of overall average performance, CoEvo-AHD achieves an average objective value of 3184.36, outperforming MATLS and S5 by approximately 2.38\% and 2.19\%, respectively, while remaining below CoCo, which obtains 3284.46. These results indicate that CoEvo-AHD is highly competitive across most problem scales, particularly on small- and medium-sized instances, where it can identify superior solutions. Meanwhile, the results on large-scale TTP instances suggest that there remains room for further improving search depth and robustness.

Table \ref{TPPResult-table} presents the experimental results on the TPP benchmark instances, where lower objective values indicate better solution quality. Unlike on TTP, CoEvo-AHD demonstrates a more pronounced overall advantage on TPP. Specifically, CoEvo-AHD achieves the best results on five instance scales, including $50\times150$, $100\times100$, $100\times150$, $150\times100$, and $150\times150$, and obtains the second-best results on three additional scales.

The advantage of CoEvo-AHD is particularly evident on larger-scale instances. For example, on $150\times100$ and $150\times150$, CoEvo-AHD achieves objective values of 1749.00 and 2201.71, respectively, substantially outperforming traditional heuristic methods. The overall average results further confirm this trend: CoEvo-AHD achieves an average objective value of 1850.77, outperforming DMD-ATA (ACO), TA, and ALNS. Compared with the second-best method, TA, which obtains 1903.35, CoEvo-AHD reduces the objective value by approximately 2.76\%; compared with DMD-ATA (ACO) and ALNS, it achieves reductions of approximately 4.11\% and 11.42\%, respectively.

These results indicate that CoEvo-AHD can effectively exploit the synergy between routing operators and selection operators in TPP, making it particularly suitable for larger instances where route selection and procurement decisions are more strongly coupled.

\begin{table}
  \caption{Average objective values on TTP benchmark. \textbf{Higher} is better. The best result is highlighted with gray background, and the second-best is underlined.}
  \label{TTPResult-table}
  \centering
  \begin{tabular}{lccccc}
    \toprule
    \multicolumn{6}{c}{\textbf{Traveling Thief Problem(TTP)}} \\
    \cmidrule(r){1-6}
    Method & \multicolumn{4}{c}{Size}&Avg($\uparrow$)\\
    \cmidrule(r){2-5}
           & TTP20 & TTP50 & TTP100 & TTP200&\\
    \midrule
    MATLS & 439.11 & 1677.72 & 3383.56 & 6940.83&3110.31\\
    S5    & 503.64 & 1614.97 & 3341.90 & 7004.35&3116.22\\
    CoCo  & 505.96 & 1728.61 & \cellcolor{gray!20}3548.50 & \cellcolor{gray!20}7354.76&\cellcolor{gray!20}3284.46\\
    CoEvo-AHD & \cellcolor{gray!20}583.10 & \cellcolor{gray!20}1730.60 & \underline{3405.19} & \underline{7018.53}&\underline{3184.36}\\
    \bottomrule
  \end{tabular}
\end{table}

\begin{table}
  \caption{Average objective values on TPP benchmark instances. \textbf{Lower} is better. The best result is highlighted with gray background, and the second-best is underlined.}
  \label{TPPResult-table}
  \centering
  \resizebox{\textwidth}{!}{%
  \begin{tabular}{lcccccccccccc}
    \toprule
    \multicolumn{13}{c}{\textbf{Traveling Purchaser Problem(TPP)}} \\
    \cmidrule(r){1-13}
    Method &\multicolumn{11}{c}{Size}&Avg($\downarrow$)\\
    \cmidrule(r){2-12}
         & $20\times 20$ & $30\times 30$ & $50\times 50$ & $50\times 100$ & $50\times 150$ & $100\times 50$ & $100\times 100$ & $100\times 150$ & $150\times 50$ & $150\times 100$&$150 \times 150$&\\
    \midrule
    DMD-ATA (ACO) & \cellcolor{gray!20}1214.33 & \cellcolor{gray!20}1393.18 & \cellcolor{gray!20}1506.63 & \cellcolor{gray!20}2263.24 & \underline{2924.01}&\cellcolor{gray!20}1180.18 &1986.71 &2649.27 &\cellcolor{gray!20}1246.73 &2240.95&2624.52&1929.98\\
    TA & 1215.71 & \underline{1410.25} & \underline{1539.94} & 2313.92 & 2924.30& \underline{1209.89} &\underline{1926.65} &\underline{2559.13} &1444.87 &\underline{1983.98}&\underline{2408.18}&\underline{1903.35}\\
    ALNS & 1215.75 & 1420.74 & 1591.43 & 2530.34 & 3300.01&1323.15 &2164.23 &3009.71 &1351.15 &2185.13&2891.11&2089.34\\
    CoEvo-AHD &\underline{1214.58} &1428.63  &1623.72  &\underline{2297.24}  &\cellcolor{gray!20}2890.13 &1245.40 &\cellcolor{gray!20}1897.73 &\cellcolor{gray!20}2505.75 &\underline{1304.61} &\cellcolor{gray!20}1749.00&\cellcolor{gray!20}2201.71&\cellcolor{gray!20}1850.77\\
    \bottomrule
  \end{tabular}%
  }
\end{table}

\subsection{Ablation Studies}
To validate the contribution of each component in CoEvo-AHD, we conduct ablation studies on the TTP50 and TPP$50\times 50$ datasets. We take the full CoEvo-AHD framework as the reference and compare it with several degraded variants.

Table \ref{Ablation-table} reports the ablation results of CoEvo-AHD on the TTP50 and TPP$50\times50$ instances. For TTP, higher objective values indicate better solution quality, whereas for TPP, lower objective values are preferred. We examine the performance changes when evolving only a single component-specific population and when removing access to the tool-augmented problem environment. Specifically, for TTP, Route/Tour-only evolution corresponds to evolving only the route-operator population, while Packing/Purchasing-only evolution corresponds to evolving only the packing-operator population. For TPP, Route/Tour-only evolution corresponds to evolving only the tour-operator population, while Packing/Purchasing-only evolution corresponds to evolving only the purchasing-operator population. The variant w/o tool-augmented environment indicates that the LLM-generated operators are not provided with access to the tool-augmented problem environment during evolution.

As shown in the table, the full CoEvo-AHD framework achieves the best performance on both tasks, reaching 1730.60 on TTP50 and 1623.72 on TPP$50\times50$. When only a single component-specific population is evolved, performance degrades substantially in all cases. For TTP, Route/Tour-only evolution and Packing/Purchasing-only evolution decrease to 1651.69 and 1680.37, respectively, indicating that optimizing either the route or packing operators alone is insufficient to fully capture the interaction between routing and packing decisions. For TPP, the degradation caused by single-population evolution is even more pronounced: the objective values of Route/Tour-only evolution and Packing/Purchasing-only evolution increase to 1803.18 and 1789.00, corresponding to deteriorations of approximately 11.05\% and 10.18\% relative to the full framework. This suggests that, in the Traveling Purchaser Problem, the coupling between tour and purchasing decisions plays a critical role in overall solution quality, and evolving only one decision component cannot yield effective cooperative improvements.

In addition, removing access to the tool-augmented problem environment also leads to performance degradation. On TTP50, w/o tool-augmented environment obtains 1634.79, corresponding to a decrease of approximately 5.54\% compared with the full CoEvo-AHD framework. Its performance is even worse than both single-population variants, suggesting that the high-frequency core operations encapsulated in the tool-augmented problem environment can substantially improve the effectiveness and reliability of LLM-generated operators. On TPP$50\times50$, w/o tool-augmented environment obtains 1671.47, corresponding to a deterioration of approximately 2.94\% relative to the full framework. Although this variant still outperforms the single-population variants, its performance drop indicates that standardized tool interfaces help the LLM avoid repeatedly implementing inefficient or error-prone problem-specific computations, thereby improving the quality of operators generated during evolution.


\begin{table}
  \caption{Ablation study of the key components in CoEvo-AHD. The best results are highlighted in \textbf{bold}.}
  \label{Ablation-table}
  \centering
  \begin{tabular}{lcc}
    \toprule
    & TTP50($\uparrow$) & TPP$50\times 50$($\downarrow$)\\
    \midrule
    \textbf{CoEvo-AHD (full)} & \textbf{1730.60} & \textbf{1623.72}\\
    \midrule
    Route/Tour-only evolution & 1651.69 & 1803.18 \\
    Packing/Purchasing-only evolution & 1680.37 & 1789.00 \\
    w/o tool-augmented environment & 1634.79 & 1671.47\\
    \bottomrule
  \end{tabular}
\end{table}

\section{Conclusion}
In this paper, we introduced CoEvo-AHD, an LLM-driven dual-population co-evolutionary framework for automated heuristic design in coupled combinatorial optimization problems. CoEvo-AHD co-evolves complementary operator populations, such as route and packing operators for the Traveling Thief Problem and tour and purchasing operators for the Traveling Purchaser Problem. Through cooperative evaluation, pairwise synergy scoring, and cross-component joint crossover, CoEvo-AHD effectively captures cross-population interactions and discovers synergistic heuristic combinations. For future work, we plan to extend CoEvo-AHD to problems with more than two coupled decision components and further improve its scalability, generalization, and on larger instances.

\bibliographystyle{unsrtnat}
\bibliography{references}

@article{romera2024funsearch,
  title={Mathematical Discoveries from Program Search with Large Language Models},
  author={Romera-Paredes, Bernardino and Barekatain, Mohammadamin and Novikov, Alexander and Balog, Matej and Kumar, M. Pawan and Dupont, Emilien and Ruiz, Francisco J. R. and Ellenberg, Jordan S. and Wang, Pengming and Fawzi, Omar and others},
  journal={Nature},
  volume={625},
  number={7995},
  pages={468--475},
  year={2024},
  publisher={Nature Publishing Group},
  doi={10.1038/s41586-023-06924-6}
}

@inproceedings{liu2024eoh,
  title     = {Evolution of Heuristics: Towards Efficient Automatic Algorithm Design Using Large Language Model},
  author    = {Liu, Fei and Tong, Xialiang and Yuan, Mingxuan and Lin, Xi and Luo, Fu and Wang, Zhenkun and Lu, Zhichao and Zhang, Qingfu},
  booktitle = {Proceedings of the 41st International Conference on Machine Learning},
  series    = {Proceedings of Machine Learning Research},
  volume    = {235},
  pages     = {32201--32223},
  year      = {2024},
  publisher = {PMLR}
}

@inproceedings{ye2024reevo,
  title     = {{ReEvo}: Large Language Models as Hyper-Heuristics with Reflective Evolution},
  author    = {Ye, Haoran and Wang, Jiarui and Cao, Zhiguang and Berto, Federico and Hua, Chuanbo and Kim, Haeyeon and Park, Jinkyoo and Song, Guojie},
  booktitle = {Advances in Neural Information Processing Systems},
  volume    = {37},
  year      = {2024},
  url       = {https://papers.nips.cc/paper_files/paper/2024/hash/4ced59d480e07d290b6f29fc8798f195-Abstract-Conference.html}
}

@inproceedings{dat2025hsevo,
  title     = {{HSEvo}: Elevating Automatic Heuristic Design with Diversity-Driven Harmony Search and Genetic Algorithm Using {LLMs}},
  author    = {Dat, Pham Vu Tuan and Doan, Long and Binh, Huynh Thi Thanh},
  booktitle = {Proceedings of the AAAI Conference on Artificial Intelligence},
  volume    = {39},
  number    = {25},
  pages     = {26931--26938},
  year      = {2025},
  doi       = {10.1609/aaai.v39i25.34898}
}

@inproceedings{zheng2025mctsahd,
  title     = {Monte Carlo Tree Search for Comprehensive Exploration in {LLM}-Based Automatic Heuristic Design},
  author    = {Zheng, Zhi and Xie, Zhuoliang and Wang, Zhenkun and Hooi, Bryan},
  booktitle = {Proceedings of the 42nd International Conference on Machine Learning},
  series    = {Proceedings of Machine Learning Research},
  volume    = {267},
  pages     = {78338--78373},
  year      = {2025},
  publisher = {PMLR},
  url       = {https://proceedings.mlr.press/v267/zheng25o.html}
}

@inproceedings{yao2025meoh,
  title     = {Multi-Objective Evolution of Heuristic Using Large Language Model},
  author    = {Yao, Shunyu and Liu, Fei and Lin, Xi and Lu, Zhichao and Wang, Zhenkun and Zhang, Qingfu},
  booktitle = {Proceedings of the AAAI Conference on Artificial Intelligence},
  volume    = {39},
  number    = {25},
  pages     = {27144--27152},
  year      = {2025},
  doi       = {10.1609/aaai.v39i25.34922}
}

@inproceedings{kiet2026motif,
  title={MOTIF: Multi-strategy Optimization via Turn-based Interactive Framework},
  author={Kiet, Nguyen Viet Tuan and Dao, Tung and Tran, Cong Dao and Binh, Huynh Thi Thanh},
  booktitle={Proceedings of the AAAI Conference on Artificial Intelligence},
  volume={40},
  number={43},
  pages={37000--37008},
  year={2026},
  doi={10.1609/aaai.v40i43.41028}
}

@inproceedings{faulkner2015approximate,
  title     = {Approximate Approaches to the Traveling Thief Problem},
  author    = {Faulkner, Hayden and Polyakovskiy, Sergey and Schultz, Tom and Wagner, Markus},
  booktitle = {Proceedings of the Genetic and Evolutionary Computation Conference},
  pages     = {385--392},
  year      = {2015}
}

@inproceedings{elyafrani2018population,
  title     = {Population-based vs. Single-solution Heuristics for the Travelling Thief Problem},
  author    = {El Yafrani, Mohamed and Ahiod, Bela{\"i}d},
  booktitle = {Proceedings of the Genetic and Evolutionary Computation Conference 2016},
  series    = {GECCO '16},
  pages     = {317--324},
  year      = {2016},
  publisher = {Association for Computing Machinery},
  address   = {New York, NY, USA},
  doi       = {10.1145/2908812.2908847},
  isbn      = {9781450342063},
  location  = {Denver, Colorado, USA}
}

@inproceedings{mei2014matls,
  title     = {Improving Efficiency of Heuristics for the Large Scale Traveling Thief Problem},
  author    = {Mei, Yi and Li, Xiaodong and Salim, Flora and Yao, Xin},
  booktitle = {Proceedings of the Asia-Pacific Conference on Simulated Evolution and Learning},
  pages     = {631--643},
  year      = {2014},
  publisher = {Springer}
}

@inproceedings{wagner2016ants,
  title     = {Stealing Items More Efficiently with Ants: A Swarm Intelligence Approach to the Travelling Thief Problem},
  author    = {Wagner, Markus},
  booktitle = {Proceedings of the International Conference on Swarm Intelligence},
  pages     = {273--281},
  year      = {2016},
  publisher = {Springer}
}

@article{wu2017exact,
  title   = {Exact Approaches for the Travelling Thief Problem},
  author  = {Wu, Junhua and Polyakovskiy, Sergey and Wagner, Markus and Neumann, Frank},
  journal = {Algorithmica},
  volume  = {82},
  pages   = {2090--2111},
  year    = {2020}
}

@article{namazi2023coco,
  title   = {Solving Travelling Thief Problems Using Coordination Based Methods},
  author  = {Namazi, Majid and Newton, M. A. Hakim and Sanderson, Conrad and Sattar, Abdul},
  journal = {Journal of Heuristics},
  volume  = {29},
  number  = {4--6},
  pages   = {487--544},
  year    = {2023},
  doi     = {10.1007/s10732-023-09518-7}
}

@article{manerba2017tppsurvey,
  title={The Traveling Purchaser Problem and Its Variants},
  author={Manerba, Daniele and Mansini, Renata and Riera-Ledesma, Jorge},
  journal={European Journal of Operational Research},
  volume={259},
  number={1},
  pages={1--18},
  year={2017},
  publisher={Elsevier},
  doi={10.1016/j.ejor.2016.12.017}
}

@article{golden1981two,
  title   = {Two Generalizations of the Traveling Salesman Problem},
  author  = {Golden, Bruce L. and Levy, Lawrence and Dahl, Robert},
  journal = {Omega},
  volume  = {9},
  number  = {4},
  pages   = {439--441},
  year    = {1981}
}

@article{ong1982approximate,
  title   = {Approximate Algorithms for the Traveling Purchaser Problem},
  author  = {Ong, H. L.},
  journal = {Operations Research Letters},
  volume  = {1},
  number  = {5},
  pages   = {201--205},
  year    = {1982}
}

@article{pearn1998improved,
  title   = {Improved Solutions for the Traveling Purchaser Problem},
  author  = {Pearn, W. L. and Chien, R. C.},
  journal = {Computers \& Operations Research},
  volume  = {25},
  number  = {11},
  pages   = {879--885},
  year    = {1998}
}

@article{boctor2003heuristics,
  title   = {Heuristics for the Traveling Purchaser Problem},
  author  = {Boctor, Fayez F. and Laporte, Gilbert and Renaud, Jacques},
  journal = {Computers \& Operations Research},
  volume  = {30},
  number  = {4},
  pages   = {491--504},
  year    = {2003},
  doi     = {10.1016/S0305-0548(02)00020-5}
}

@article{voss1996tabu,
  title   = {Dynamic Tabu Search Strategies for the Traveling Purchaser Problem},
  author  = {Vo{\ss}, Stefan},
  journal = {Annals of Operations Research},
  volume  = {63},
  pages   = {253--275},
  year    = {1996}
}

@article{laporte2003branchcut,
  title   = {A Branch-and-Cut Algorithm for the Undirected Traveling Purchaser Problem},
  author  = {Laporte, Gilbert and Riera-Ledesma, Jorge and Salazar-Gonz{\'a}lez, Juan-Jos{\'e}},
  journal = {Operations Research},
  volume  = {51},
  number  = {6},
  pages   = {940--951},
  year    = {2003}
}

@article{bontoux2008aco,
  title   = {Ant Colony Optimization for the Traveling Purchaser Problem},
  author  = {Bontoux, Boris and Feillet, Dominique},
  journal = {Computers \& Operations Research},
  volume  = {35},
  number  = {2},
  pages   = {628--637},
  year    = {2008},
  doi     = {10.1016/j.cor.2006.03.023}
}

@article{goldbarg2009ta,
  title   = {Transgenetic Algorithm for the Traveling Purchaser Problem},
  author  = {Goldbarg, Marco C. and Bagi, Ligia B. and Goldbarg, Elizabeth F. G.},
  journal = {European Journal of Operational Research},
  volume  = {199},
  number  = {1},
  pages   = {36--45},
  year    = {2009},
  doi     = {10.1016/j.ejor.2008.10.027}
}

@article{yuan2024drl,
  title={Deep Reinforcement Learning for Traveling Purchaser Problems},
  author={Yuan, Haofeng and Zhu, Rongping and Yang, Wanlu and Song, Shiji and You, Keyou and Fan, Wei and Chen, C. L. Philip},
  journal={IEEE Transactions on Emerging Topics in Computational Intelligence},
  volume={10},
  number={1},
  pages={425--439},
  year={2026},
  publisher={IEEE},
  doi={10.1109/TETCI.2025.3581113}
}

@article{kucukoglu2024tpppf,
  title   = {The Traveling Purchaser Problem for Perishable Foods},
  author  = {Kucukoglu, Ilker and Vansteenwegen, Pieter and Cattrysse, Dirk},
  journal = {Computers \& Industrial Engineering},
  volume  = {195},
  pages   = {110424},
  year    = {2024}
}

@article{kapancioglu2025ils,
  title   = {An Iterated Local Search Algorithm for the Traveling Purchaser Problem},
  author  = {Kapancioglu, Tom{\'a}s and Bernardino, Raquel},
  journal = {European Journal of Operational Research},
  volume  = {324},
  number  = {3},
  pages   = {759--772},
  year    = {2025},
  doi     = {10.1016/j.ejor.2025.02.024}
}

@article{el2018efficiently,
  title={Efficiently Solving the Traveling Thief Problem Using Hill Climbing and Simulated Annealing},
  author={El Yafrani, Mohamed and Ahiod, Bela{\"i}d},
  journal={Information Sciences},
  volume={432},
  pages={231--244},
  year={2018},
  publisher={Elsevier},
  doi={10.1016/j.ins.2017.12.011}
}

@misc{RieraLedesma2012TPPLIB,
  author       = {Riera-Ledesma, Jorge},
  title        = {{TPPLIB}},
  year         = {2012},
  howpublished = {\url{https://jriera.webs.ull.es/TPP.htm}},
  note         = {Benchmark instances for the Traveling Purchaser Problem, accessed 2026-05-07}
}

@inproceedings{yang2024large,
  title={{Large Language Models as Optimizers}},
  author={Yang, Chengrun and Wang, Xuezhi and Lu, Yifeng and Liu, Hanxiao and Le, Quoc V. and Zhou, Denny and Chen, Xinyun},
  booktitle={The Twelfth International Conference on Learning Representations},
  year={2024}
}

@article{zhao2026glns,
  title={{G-LNS}: Generative Large Neighborhood Search for {LLM}-Based Automatic Heuristic Design},
  author={Zhao, Baoyun and Wang, He and Zeng, Liang},
  journal={arXiv preprint arXiv:2602.08253},
  year={2026},
  eprint={2602.08253},
  archivePrefix={arXiv},
  primaryClass={cs.AI}
}

@article{qiu2026e2oc,
  title={Evolving Interdependent Operators with Large Language Models for Multi-Objective Combinatorial Optimization},
  author={Qiu, Junhao and Chen, Xin and Ge, Liang and Lin, Liyong and Lu, Zhichao and Zhang, Qingfu},
  journal={arXiv preprint arXiv:2601.17899},
  year={2026},
  eprint={2601.17899},
  archivePrefix={arXiv},
  primaryClass={cs.AI}
}

@inproceedings{liu2026eohs,
  title={{EoH-S}: Evolution of Heuristic Set Using {LLMs} for Automated Heuristic Design},
  author={Liu, Fei and Liu, Yilu and Zhang, Qingfu and Tong, Xialiang and Yuan, Mingxuan},
  booktitle={Proceedings of the AAAI Conference on Artificial Intelligence},
  volume={40},
  number={43},
  pages={37090--37098},
  year={2026},
  doi={10.1609/aaai.v40i43.41038}
}

@inproceedings{bonyadi2013ttp,
  title={The Travelling Thief Problem: The First Step in the Transition from Theoretical Problems to Realistic Problems},
  author={Bonyadi, Mohammad Reza and Michalewicz, Zbigniew and Barone, Luigi},
  booktitle={2013 IEEE Congress on Evolutionary Computation},
  pages={1037--1044},
  year={2013},
  organization={IEEE},
  doi={10.1109/CEC.2013.6557681}
}

@inproceedings{polyakovskiy2014benchmark,
  title={A Comprehensive Benchmark Set and Heuristics for the Traveling Thief Problem},
  author={Polyakovskiy, Sergey and Bonyadi, Mohammad Reza and Wagner, Markus and Michalewicz, Zbigniew and Neumann, Frank},
  booktitle={GECCO 2014: Proceedings of the 2014 Genetic and Evolutionary Computation Conference},
  pages={477--484},
  year={2014},
  publisher={Association for Computing Machinery},
  doi={10.1145/2576768.2598249},
  isbn={9781450326629}
}







\appendix
%
%

\emergencystretch=2em
\definecolor{coevoNavy}{RGB}{0,0,128}
\definecolor{coevoCodeGreen}{RGB}{28,172,0}
\definecolor{coevoPurple}{RGB}{160,32,240}

\lstdefinestyle{coevoPython}{%
    language=Python,
    basicstyle=\ttfamily\footnotesize,
    keywordstyle=\color{blue}\bfseries,
    commentstyle=\color{gray},
    stringstyle=\color{red},
    breaklines=true,
    breakatwhitespace=false,
    columns=fullflexible,
    keepspaces=true,
    showstringspaces=false,
    frame=single,
    rulecolor=\color{coevoNavy},
    xleftmargin=0.5em,
    xrightmargin=0.5em,
    aboveskip=0.75em,
    belowskip=0.75em
}

\setlength{\abovedisplayskip}{5pt plus 2pt minus 2pt}
\setlength{\belowdisplayskip}{5pt plus 2pt minus 2pt}
\setlength{\abovedisplayshortskip}{3pt plus 1pt minus 1pt}
\setlength{\belowdisplayshortskip}{3pt plus 1pt minus 1pt}

\newtcolorbox{operatorbox}[1]{%
    enhanced,
    breakable,
    colback=white,
    colframe=coevoNavy,
    boxrule=1pt,
    arc=2mm,
    top=2.5mm,
    bottom=2.5mm,
    left=3mm,
    right=3mm,
    before skip=1em,
    after skip=1em,
    colbacktitle=coevoNavy,
    coltitle=white,
    fonttitle=\bfseries,
    title={#1}
}

\section{Additional Related Work and Positioning}
\label{app:related_work}

This appendix provides additional discussion on how CoEvo-AHD differs from
existing LLM-driven automated heuristic design methods and multi-operator
optimization frameworks.

Existing LLM-AHD methods mainly generate and evolve heuristic functions,
construction rules, local-search operators, or algorithmic strategies. Their
operator complementarity is usually defined at the level of algorithmic roles,
such as construction versus improvement, destroy versus repair, or different
search policies. In contrast, CoEvo-AHD studies complementarity induced by the
structure of the optimization problem itself. Specifically, each operator
population is tied to one decision component of the complete solution, and an
operator's utility is evaluated through its interaction with the other component.

For TTP, the two decision components are routing and packing. A route move
changes the transportation cost of selected items, while a packing decision
changes the effective cost landscape of route modifications. For TPP, the two
decision components are market-route selection and purchasing. A route
modification changes the feasible set of purchasing choices, while a purchasing
decision changes the marginal value of visiting or removing markets. Therefore,
the quality of a component operator cannot be reliably assessed in isolation.

CoEvo-AHD is designed around this relational nature of operator utility. It
maintains component-specific populations, evaluates selected operator pairs
through complete-solution trajectories, records pair-level collaboration scores,
and uses cross-component joint crossover to generate mutually compatible
operators. These mechanisms distinguish CoEvo-AHD from methods that evolve a
single heuristic, a homogeneous operator set, or operators tied only to search
workflow roles.

\section{Problem Formulations}
\label{app:problem_formulations}

This section provides the mathematical formulations of the two coupled
combinatorial optimization problems used to instantiate CoEvo-AHD. The Traveling
Purchaser Problem (TPP) couples market selection, routing, and purchasing
decisions, whereas the Traveling Thief Problem (TTP) couples routing and packing
decisions through a load-dependent travel speed.

\subsection{Traveling Purchaser Problem}
\label{app:tpp_formulation}

The Traveling Purchaser Problem (TPP) generalizes the traveling salesman problem
by jointly considering market selection, route construction, and product
purchasing. Let $0$ denote the depot, $N=\{1,\ldots,n\}$ denote the set of
markets, $K=\{1,\ldots,m\}$ denote the set of products, and $V=N\cup\{0\}$.
For each arc $(i,j)$ with $i,j\in V$ and $i\neq j$, let $c_{ij}$ be the travel
cost. For market $i\in N$ and product $k\in K$, let $p_{ik}$ be the unit
purchase price, $u_{ik}$ be the available supply, and $d_k$ be the demand.

We define binary routing variables $x_{ij}$, binary market-visit variables
$y_i$, and continuous purchasing variables $q_{ik}$ as follows:
\[
x_{ij}=
\begin{cases}
1, & \text{if the route travels directly from } i \text{ to } j,\\
0, & \text{otherwise,}
\end{cases}
\quad
y_i=
\begin{cases}
1, & \text{if market } i \text{ is visited,}\\
0, & \text{otherwise,}
\end{cases}
\]
and
\[
q_{ik}\geq 0,
\quad i\in N,\ k\in K.
\]

A standard mixed-integer formulation of TPP is:
\begin{align}
\min \quad
& \sum_{\substack{i,j\in V\\i\neq j}} c_{ij}x_{ij}
+
\sum_{i\in N}\sum_{k\in K}p_{ik}q_{ik}
\label{eq:tpp_obj}
\\
\text{s.t.}\quad
& \sum_{j\in N}x_{0j}=1,
\label{eq:tpp_depot_out}
\\
& \sum_{i\in N}x_{i0}=1,
\label{eq:tpp_depot_in}
\\
& \sum_{\substack{j\in V\\j\neq i}}x_{ij}=y_i,
&& \forall i\in N,
\label{eq:tpp_flow_out}
\\
& \sum_{\substack{j\in V\\j\neq i}}x_{ji}=y_i,
&& \forall i\in N,
\label{eq:tpp_flow_in}
\\
& \sum_{i\in N}q_{ik}=d_k,
&& \forall k\in K,
\label{eq:tpp_demand}
\\
& 0\leq q_{ik}\leq u_{ik}y_i,
&& \forall i\in N,\ k\in K,
\label{eq:tpp_availability}
\\
& \sum_{\substack{i,j\in S\\i\neq j}}x_{ij}
\leq |S|-1,
&& \forall S\subseteq N,\ |S|\geq 2,
\label{eq:tpp_subtour}
\\
& x_{ij}\in\{0,1\},
&& \forall i,j\in V,\ i\neq j,
\label{eq:tpp_binary_x}
\\
& y_i\in\{0,1\},
&& \forall i\in N.
\label{eq:tpp_binary_y}
\end{align}

Constraints~\eqref{eq:tpp_depot_out}--\eqref{eq:tpp_depot_in} ensure that the
route starts and ends at the depot. Constraints~\eqref{eq:tpp_flow_out} and
\eqref{eq:tpp_flow_in} link market visits with routing decisions. Constraint
\eqref{eq:tpp_demand} enforces exact demand satisfaction, while
Constraint~\eqref{eq:tpp_availability} ensures that products can only be
purchased from visited markets and cannot exceed available supply. Constraint
\eqref{eq:tpp_subtour} eliminates cycles disconnected from the depot. Since
unvisited markets have zero in-degree and out-degree by
Constraints~\eqref{eq:tpp_flow_out}--\eqref{eq:tpp_flow_in}, the subtour
constraint remains valid for selected-market routing.

\subsection{Traveling Thief Problem}
\label{app:ttp_formulation}

The Traveling Thief Problem (TTP) couples the Traveling Salesman Problem with
the Knapsack Problem. Given a set of cities, a set of items, a knapsack capacity,
a maximum speed, a minimum speed, and a renting rate, the decision maker needs
to determine a closed tour visiting all cities and decide which items to pick,
so as to maximize the difference between collected item profit and travel
renting cost.

Let $0$ denote the starting city, $N=\{1,\ldots,n\}$ denote the remaining cities,
$V=N\cup\{0\}$ denote the set of all cities, and $J=\{1,\ldots,m\}$ denote the
set of items. For any $i,j\in V$ with $i\neq j$, let $c_{ij}$ be the travel
distance from city $i$ to city $j$. For item $k\in J$, let $p_k$ denote its
profit, $w_k$ denote its weight, and $a_k\in N$ denote the city where it is
located. The knapsack capacity is $W$, the maximum and minimum speeds are
$v_{\max}$ and $v_{\min}$, respectively, and the renting rate is $R$.

The routing decision is represented by a city permutation
\begin{equation}
    \pi=(\pi_0,\pi_1,\ldots,\pi_n),
\end{equation}
where $\pi_0=0$, $\{\pi_1,\ldots,\pi_n\}=N$, and $\pi_{n+1}=\pi_0$ denotes
the return to the starting city. The packing decision is represented by
\begin{equation}
    z_k\in\{0,1\},
    \quad k\in J.
\end{equation}

For route position $t$, the accumulated knapsack load when leaving city $\pi_t$
is
\begin{equation}
    L_t(\pi,z)
    =
    \sum_{k\in J:\, a_k\in\{\pi_0,\pi_1,\ldots,\pi_t\}}
    w_k z_k,
    \quad t=0,1,\ldots,n .
\end{equation}
The travel speed decreases linearly with the accumulated load:
\begin{equation}
    v_t(\pi,z)
    =
    v_{\max}
    -
    \left(v_{\max}-v_{\min}\right)
    \frac{L_t(\pi,z)}{W}.
\end{equation}

The TTP can be written as
\begin{align}
\max_{\pi,z}\quad
&
\sum_{k\in J} p_k z_k
-
R
\sum_{t=0}^{n}
\frac{c_{\pi_t,\pi_{t+1}}}{v_t(\pi,z)}
\label{eq:ttp_obj}
\\
\text{s.t.}\quad
& \pi_0 = 0,
\label{eq:ttp_depot}
\\
& \{\pi_1,\ldots,\pi_n\}=N,
\label{eq:ttp_permutation}
\\
& \sum_{k\in J} w_k z_k \leq W,
\label{eq:ttp_capacity}
\\
& z_k\in\{0,1\},
&& \forall k\in J .
\label{eq:ttp_binary_z}
\end{align}

The objective in Eq.~\eqref{eq:ttp_obj} maximizes item profit minus route
renting cost. The travel-time term depends on both the route permutation $\pi$
and the packing plan $z$: the route determines the remaining distance over
which each picked item is carried, while the packing plan affects travel speed
through accumulated load.

\subsection{Unified Utility Convention}
\label{app:utility_convention}

Since TPP is a minimization problem and TTP is a maximization problem, we use a
unified utility function $u(\cdot)$ in the algorithmic description:
\[
u(x\mid\mathcal{I})=
\begin{cases}
-F_{\mathrm{TPP}}(x\mid\mathcal{I}), & \text{for TPP},\\
F_{\mathrm{TTP}}(x\mid\mathcal{I}), & \text{for TTP}.
\end{cases}
\]
Under this convention, a larger utility value always indicates a better
solution. All acceptance rules, rewards, and bandit updates in the generic
CoEvo-AHD framework are written in terms of $u(\cdot)$.


\section{CoEvo-AHD Implementation and Problem-specific Instantiation}
\label{app:implementation_instantiation}
\label{app:problem_instantiation}
\label{app:general_algorithm}

This section combines the implementation-level description of CoEvo-AHD with
its problem-specific instantiations on TPP and TTP. The goal is to avoid
repeating the same reconstruction, shared-information, and validation mechanisms
in separate appendix sections. The section first defines the unified component
interface, then explains the two instantiations, and finally summarizes the
common evolution procedure.

\subsection{Unified Operator Interface}
\label{app:operator_interface}

For a bi-component optimization problem, a complete solution is written as
\(x=(x^A,x^B)\), where \(x^A\) and \(x^B\) denote two coupled decision
components. CoEvo-AHD maintains two component-specific operator populations,
\(\mathcal{P}^{A}=\{o_1^A,\ldots,o_N^A\}\) and
\(\mathcal{P}^{B}=\{o_1^B,\ldots,o_N^B\}\). Each operator returns only a
candidate for its own component:
\[
\widetilde{x}^{c}=o_i^c(x^A,x^B,\mathcal{I},\mathcal{M}),
\qquad c\in\{A,B\}.
\]

The implementation uses the following concrete function signatures:
\begin{lstlisting}[style=coevoPython]
# TPP
tour_operator(tour, purchasing_plan, problem_data)
purchase_operator(tour, purchasing_plan, problem_data)

# TTP
route_operator(route, pack_plan, problem_data)
pack_operator(route, pack_plan, problem_data)
\end{lstlisting}
The shared information space \(\mathcal{M}\) is implemented as the mutable
dictionary \texttt{shared\_info} inside \texttt{problem\_data}. Operators may write soft
cross-component suggestions into this dictionary, but no suggestion is accepted
without reconstruction, feasibility checking, and complete-objective evaluation.

\subsection{Instantiation on TPP}
\label{app:tpp_instantiation}

\subsubsection{Solution representation, evaluation, and reconstruction}
\label{app:tpp_eval_reconstruct}

For TPP, the complete solution is decomposed as \(x=(x^T,x^P)\), where
\(x^T\) is a route over a selected subset of markets and \(x^P\) is a
purchasing plan represented as product-indexed purchase records. A route must
contain valid city indices and include the depot, but it is not required to
visit all markets.

The TPP environment exposes
\[
\mathcal{E}_{\mathrm{TPP}}
=\{\mathrm{Evaluate},\mathrm{GreedyPurchase},
\mathrm{CityValueAnalysis},\mathrm{DropCityEval}\}.
\]
\(\mathrm{Evaluate}\) computes the complete TPP objective or the internal
penalty-augmented search score. \(\mathrm{GreedyPurchase}\) reconstructs the
cheapest purchasing plan for a fixed route. \(\mathrm{CityValueAnalysis}\) and
\(\mathrm{DropCityEval}\) provide market-level signals for insertion, removal,
and simplification.

During evaluation, a tour proposal is first normalized and validated. Its
purchasing plan is then reconstructed by \(\mathrm{GreedyPurchase}\), and the
complete solution is accepted only if the objective decreases. A purchase
proposal is evaluated on the current route. If it improves the complete
solution, it is accepted; otherwise, the previous solution is retained.

For intermediate infeasible purchasing plans, the implementation uses the
penalty-augmented score
\[
\Phi_{\mathrm{TPP}}(x^T,x^P)
= C_{\mathrm{travel}}(x^T)+C_{\mathrm{purchase}}(x^P)
+\rho\sum_{k\in K}\max\left(0,d_k-\sum_{i\in N}q_{ik}\right)
+\Psi_{\mathrm{invalid}}(x^T,x^P).
\]
This score is used only during search and repair. Reported final objective
values are computed from validated solutions using the unified TPP evaluator.

\subsubsection{Shared messages and post-processing}
\label{app:tpp_messages}
\label{app:tpp_postprocess}

For TPP, purchase operators may write purchase-guided route suggestions such as
suggested tours, candidate tour lists, add-city lists, drop-city lists, and swap
moves. These fields are decoded into a bounded set of candidate tours. Complete
route proposals are normalized by removing duplicate cities and ensuring that
the depot is present. Add or insert messages are converted into tours by
cheapest insertion, drop messages produce compact removal candidates, and swap
messages remove one visited market while inserting one proposed market.

Each decoded route candidate is validated, completed by
\(\mathrm{GreedyPurchase}\), and evaluated as a complete TPP solution. If the
current solution is already feasible, the accepted shared-information candidate
must also be feasible. If the current solution is infeasible, a shared candidate
may be accepted as a repair move when it improves the penalty-augmented score.

After an accepted route update, purchase update, or shared-information move,
CoEvo-AHD applies a city-drop post-processing step. Candidate market removals
are tested by reconstructing the purchasing plan and re-evaluating the complete
solution. A removal is accepted only when demand satisfaction is preserved and
the original TPP objective decreases.

\subsection{Instantiation on TTP}
\label{app:ttp_instantiation}

\subsubsection{Solution representation, evaluation, and reconstruction}
\label{app:ttp_eval_pack}

For TTP, the complete solution is decomposed as \(x=(x^R,x^B)\), where
\(x^R\) is a full city permutation and \(x^B\) is a binary packing vector. The
implementation exposes
\[
\mathcal{E}_{\mathrm{TTP}}
=\{\mathrm{Evaluate},\mathrm{Fast2OptDelta},\mathrm{GreedyPack}\}.
\]
\(\mathrm{Evaluate}\) computes the complete TTP objective,
\(\mathrm{Fast2OptDelta}\) evaluates a 2-opt route move under the current
packing plan, and \(\mathrm{GreedyPack}\) constructs a route-aware packing plan.

Given a route \(x^R\), the fallback route-aware packing rule scores item \(k\),
located at city \(a_k\), by
\[
s_k(\alpha)=
\frac{(p_k/\max(w_k,1))^\alpha}
{\max(D_{\mathrm{suffix}}(a_k;x^R),10^{-9})},
\]
where \(D_{\mathrm{suffix}}(a_k;x^R)\) is the remaining route distance after
the item city. Items are inserted in descending score order while respecting the
capacity constraint.

During evaluation, a route proposal must be a valid full permutation and is
accepted only if it improves the complete TTP objective. A packing proposal must
be a binary vector satisfying the knapsack capacity and is accepted only if it
improves the complete objective.

\subsubsection{Shared messages and post-processing}
\label{app:ttp_messages}
\label{app:ttp_postprocess}

For TTP, route operators may write route-side structural signals such as route
positions, suffix distances, city-load profiles, loaded-edge information, and
acceleration segments. Packing operators may read these fields and may write
load, value, and position signals, including city burden scores, net pick-value
signals, early/late pressure indicators, critical items, heavy cities, valuable
cities, and compact route-reordering hints.

Because TTP must visit every city, shared route suggestions never add or drop
cities. The evaluator interprets them as soft reordering signals and constructs
full-permutation route candidates. For each candidate route, the evaluator
applies distance and load-order guards, reconstructs route-aware packing
candidates, evaluates the complete TTP objective, and accepts only the best
improving feasible route--pack solution.

After an accepted route update, packing update, or pack-guided shared candidate,
CoEvo-AHD applies an item-flip post-processing step. A flip is tested only if it
preserves knapsack capacity and is retained only if it improves the exact TTP
objective.

\subsection{Comparison of the Two Instantiations}
\label{app:instantiation_comparison}

\begin{table}[H]
\centering
\caption{Comparison of the TPP and TTP instantiations of CoEvo-AHD.}
\label{tab:instantiation_comparison}
\small
\begin{tabularx}{\textwidth}{p{3.1cm}XX}
\toprule
Aspect & TPP & TTP \\
\midrule
Decision components & Market-route selection and purchasing & Full route permutation and binary packing \\
Operator pair & Tour operator and purchase operator & Route operator and pack operator \\
Evaluation direction & Minimization & Maximization \\
Reconstruction & Recompute purchasing for a route by greedy purchase & Recompute or refine route-aware packing for a route \\
Shared information & Add, drop, swap, and candidate-tour suggestions & Route-position, load-profile, item-value, and reordering signals \\
Post-processing & City-drop improvement & Item-flip improvement \\
\bottomrule
\end{tabularx}
\end{table}

\subsection{Component-wise Thompson Sampling and Pair Score Recording}
\label{app:operator_selection}

In each collaborative evaluation round, CoEvo-AHD selects one operator from each
component population. The implementation uses component-wise local Thompson
sampling rather than directly sampling over all possible operator pairs. A
lightweight local bandit is initialized inside each per-instance evaluation
worker. For each operator \(o\), it maintains the number of observations, reward
sum, squared reward sum, and reward history.

If an operator has fewer than two observations, it is sampled from a wide
exploratory Gaussian. Otherwise, the local sampler draws
\[
\theta_o\sim\mathcal{N}\left(\mu_o,\frac{\nu_o}{n_o}\right),
\]
where \(\mu_o\), \(\nu_o\), and \(n_o\) are its empirical mean, empirical
variance, and observation count. The two component operators are selected
independently by this rule and then evaluated as a pair. Pair-level
collaboration scores are recorded after evaluation and used for best-pair
tracking and cross-component joint crossover.

\subsection{Candidate Reconstruction and Acceptance Rule}
\label{app:repair_reconstruct}

Because the two components are coupled, every component-level proposal is
reconstructed into a complete solution before acceptance. The problem-specific
rules are those described in the TPP and TTP instantiations above. In general,
the evaluator follows
\[
x_{t+1}
=
\begin{cases}
\bar{x}, & \text{if }\bar{x}\text{ is feasible and improves the current complete solution},\\
x_t, & \text{otherwise.}
\end{cases}
\]
Thus, component outputs and shared messages act as candidate generators rather
than direct replacements of the current solution.

\subsection{Reward and Credit Assignment}
\label{app:reward_credit}

The reward rule is instantiated differently for the two objectives. For TPP,
which is a minimization problem, let \(F_0\) be the initial objective value of a
restart, \(F_{\mathrm{best}}\) the best objective found in that restart, and
\(F_{\mathrm{base}}\) the objective value of the deterministic baseline. The
reward is
\[
r_{\mathrm{TPP}}
=
\frac{F_0-F_{\mathrm{best}}}{\max(|F_0|,1)}
+0.5\frac{F_{\mathrm{base}}-F_{\mathrm{best}}}{\max(|F_{\mathrm{base}}|,1)}.
\]
This reward is added to the selected tour operator, purchase operator, and their
pair.

For TTP, which is a maximization problem, the evaluator decomposes the raw gain
into direct improvement and pack-guided proposal improvement:
\[
\begin{gathered}
g_{\mathrm{raw}}=F_{\mathrm{best}}-F_0,\qquad
\bar{g}_{\mathrm{prop}}=
\min\{\max(0,g_{\mathrm{prop}}),\max(0,g_{\mathrm{raw}})\},\\
g_{\mathrm{dir}}=g_{\mathrm{raw}}-\bar{g}_{\mathrm{prop}},\qquad
r_R=g_{\mathrm{dir}}+\lambda_R\bar{g}_{\mathrm{prop}},\quad
r_B=g_{\mathrm{dir}}+\lambda_B\bar{g}_{\mathrm{prop}},\quad
r_{RB}=g_{\mathrm{raw}}.
\end{gathered}
\]
In the implementation, \(\lambda_R=0.50\) and \(\lambda_B=0.80\). The route and
pack local Thompson samplers are updated with \(r_R\) and \(r_B\), respectively,
while the pair-level collaboration score is updated with \(r_{RB}\).

\subsection{Population Management and LLM-driven Evolution}
\label{app:population_evolution}

At the end of each generation, each component population is ranked according to
its individual operator score. For component \(c\in\{A,B\}\), the framework keeps
\(\mathrm{Top}_{N-M}(\mathcal{P}^{c};F^c)\) and removes the lowest-ranked
\(M\) operators. Reward histories and pair-level statistics associated with
removed operators are also removed.

The vacancies are filled by LLM-generated operators through three actions:
mutation, homogeneous crossover, and cross-component joint crossover. Mutation
rewrites a single parent operator. Homogeneous crossover recombines two
operators from the same component population. Cross-component joint crossover
selects the highest-scoring operator pair and asks the LLM to rewrite it as a
coordinated pair. Generated operators must use the required component
interfaces and pass the validation pipeline before insertion.

\subsection{Overall Pseudocode}
\label{app:overall_pseudocode}

\begin{algorithm}[H]
\caption{CoEvo-AHD Implementation-Level Procedure}
\label{alg:coevo_ahd_appendix}
\begin{algorithmic}[1]
\Require Training instances $\mathcal{I}_{\mathrm{train}}$, generations $G$,
population size $N$, pruning size $M$, evaluation rounds $R$, search iterations $B$
\Ensure Evolved operator populations $\mathcal{P}^{A},\mathcal{P}^{B}$ and best evolved pair $(o_{i^\star}^{A},o_{j^\star}^{B})$

\State Initialize component populations $\mathcal{P}^{A}$ and $\mathcal{P}^{B}$
\State Validate generated operators before insertion
\State Initialize global operator statistics and pair scores

\For{$g=1,\ldots,G$}
    \For{each training instance $\mathcal{I}$ in parallel}
        \State Initialize a local Thompson sampler for all valid operators
        \For{$r=1,\ldots,R$}
            \State Select $o_i^A$ and $o_j^B$ by component-wise Thompson sampling
            \State Construct an initial complete solution $x=(x^A,x^B)$
            \State Initialize or clear \texttt{problem\_data['shared\_info']}
            \For{$b=1,\ldots,B$}
                \State Execute $o_i^A$, validate its component output, reconstruct a complete solution, and accept if improved
                \State Execute $o_j^B$, validate its component output, reconstruct a complete solution, and accept if improved
                \State Decode shared-information signals into candidate complete solutions
                \State Validate, reconstruct, evaluate, and accept the best improving candidate if any
            \EndFor
            \State Compute rewards for the selected component operators and their pair
            \State Update local Thompson samplers and pair statistics
        \EndFor
        \State Return local rewards, pair scores, best objectives, and diagnostics
    \EndFor

    \State Aggregate rewards, histories, pair scores, and diagnostics across instances
    \State Update global operator scores and pair-level collaboration scores
    \State Update the best evolved operator pair if validation performance improves
    \State Prune the lowest-ranked $M$ operators from each component population
    \State Generate replacements by mutation, homogeneous crossover, or joint crossover
    \State Validate generated operators before insertion
\EndFor

\State \Return $\mathcal{P}^{A}$, $\mathcal{P}^{B}$, and $(o_{i^\star}^{A},o_{j^\star}^{B})$
\end{algorithmic}
\end{algorithm}

\section{Prompt Templates and Operator Validation}
\label{app:prompts_validation}

\subsection{Prompt Templates}
\label{app:prompt_templates}

This section reports compact prompt skeletons used for LLM-driven operator
generation and evolution. In the implementation, these skeletons are
instantiated separately for TPP and TTP with problem-specific function
signatures, FastEnv tools, and shared-information keys. All generated operators
are required to follow the component interface defined in
Appendix~\ref{app:operator_interface}. The raw LLM output is not inserted into
the population directly. Instead, it must pass the validation pipeline in
Appendix~\ref{app:operator_validation}.

\begin{operatorbox}{Prompt for Operator Initialization}
\small

You are an expert in heuristic optimization for combinatorial problems.
Your task is to design a new \textbf{\{component\_type\} Operator} for
\textbf{\{problem\_type\}}.

\textbf{Problem Description:} \{task\_description\}

\textbf{Existing \{component\_type\} Operators:} \{existing\_operators\}

\textbf{Auxiliary Protocol:} \{auxiliary\_protocol\}

\textbf{Requirements:}
\begin{enumerate}[leftmargin=*, noitemsep, topsep=0pt]
    \item First, describe the new algorithm and its main steps in one sentence.
    The description must be inside braces \texttt{\{...\}}.
    \item The logic should be different from the provided reference operators
    to improve population diversity.
    \item Implement it in Python using the required function signature:
    \texttt{\{function\_signature\}}.
    \item The function must return a feasible output for the corresponding
    component.
    \item Only use standard Python libraries and NumPy.
    \item If helper functions are needed, define them inside the main function.
    \item Use the exposed environment tools when available.
    \item Wrap code in \texttt{```python ... ```}.
    \item Do not provide additional explanations outside the required output.
\end{enumerate}
\end{operatorbox}

\begin{operatorbox}{Prompt for Mutation}
\small

You are an algorithm optimizer. We have a
\textbf{\{component\_type\} Operator} for \textbf{\{problem\_type\}}.

\textbf{Problem Description:} \{task\_description\}

\textbf{Strategy:} \{advice\}

\textbf{Current Code:} \{operator\_code\}

\textbf{Optional Execution Profile:} \{profiling\_report\}

\textbf{Task:}
Refine and improve this operator code according to the strategy.

\textbf{Requirements:}
\begin{enumerate}[leftmargin=*, noitemsep, topsep=0pt]
    \item Keep the required function signature unchanged:
    \texttt{\{function\_signature\}}.
    \item The function must return a feasible output for the corresponding
    component.
    \item Only use standard Python libraries and NumPy.
    \item Define helper functions inside the main function if needed.
    \item Use exposed environment tools when available.
    \item Wrap code in \texttt{```python ... ```}.
    \item Do not provide additional explanations outside the required output.
\end{enumerate}
\end{operatorbox}

\begin{operatorbox}{Prompt for Homogeneous Crossover}
\small

You are an expert in heuristic optimization. Create a new
\textbf{\{component\_type\} Operator} for \textbf{\{problem\_type\}} by combining
ideas from two parent operators of the same component type.

\textbf{Problem Description:} \{task\_description\}

\textbf{Parent Operator 1:} \{parent1\_code\}

\textbf{Parent Operator 2:} \{parent2\_code\}

\textbf{Task:}
Identify useful mechanisms from both parents and synthesize a new operator.
The new operator should not simply copy either parent.

\textbf{Requirements:}
\begin{enumerate}[leftmargin=*, noitemsep, topsep=0pt]
    \item First, describe the new algorithm and its main steps in one sentence.
    The description must be inside braces \texttt{\{...\}}.
    \item Implement the new operator using the required function signature:
    \texttt{\{function\_signature\}}.
    \item The function must return a feasible output for the corresponding
    component.
    \item Only use standard Python libraries and NumPy.
    \item Define helper functions inside the main function if needed.
    \item Use exposed environment tools when available.
    \item Wrap code in \texttt{```python ... ```}.
    \item Do not provide additional explanations outside the required output.
\end{enumerate}
\end{operatorbox}

\begin{operatorbox}{Prompt for Cross-Component Joint Crossover}
\small

You are an expert in heuristic optimization. We use cross-component joint
crossover to co-evolve two heterogeneous operators for
\textbf{\{problem\_type\}}.

\textbf{Component A Description:} \{task\_description\_A\}

\textbf{Component B Description:} \{task\_description\_B\}

\textbf{Best-performing Operator Pair:}

\textbf{Operator A Code:} \{operator\_A\_code\}

\textbf{Operator B Code:} \{operator\_B\_code\}

\textbf{Synergy Objective:} \{synergy\_rule\}

\textbf{Task:}
Rewrite this pair as a coordinated entity. The two new operators should be
mutually compatible rather than independently optimized.

\textbf{Requirements:}
\begin{enumerate}[leftmargin=*, noitemsep, topsep=0pt]
    \item First, describe the joint design idea and its main steps in one
    sentence. The description must be inside braces \texttt{\{...\}}.
    \item Return one code block containing both generated operators.
    \item Use the required function signatures:
    \texttt{\{function\_signature\_A\}} and
    \texttt{\{function\_signature\_B\}}.
    \item Each operator must return a feasible output for its corresponding
    component.
    \item Only use standard Python libraries and NumPy.
    \item Define helper functions inside each main function if needed.
    \item Use shared information and exposed environment tools when useful.
    \item Wrap code in \texttt{```python ... ```}.
    \item Do not provide additional explanations outside the required output.
\end{enumerate}
\end{operatorbox}


\subsection{Validation Pipeline for LLM-Generated Operators}
\label{app:operator_validation}

CoEvo-AHD does not directly trust raw LLM-generated code. The implementation
validates generated code before insertion and again during per-instance
evaluation.

\paragraph{Code extraction and registration.}
Generated text must contain a Python code block. The code is executed in a
restricted namespace containing standard Python utilities, NumPy, and the
problem-specific FastEnv class when enabled. The framework then searches for
the required function name: \texttt{tour\_operator} or
\texttt{purchase\_operator} for TPP, and \texttt{route\_operator} or
\texttt{pack\_operator} for TTP. If the exact function name is absent, the
implementation attempts to locate a callable with a compatible signature.

\paragraph{TPP validation.}
For TPP, operators are invoked with a timeout during evaluation. A tour proposal
must be a valid list of city indices containing the depot and no duplicate
cities. A purchasing proposal must be a dictionary mapping product indices to
lists of \((\text{city},\text{quantity})\) entries. Invalid outputs, exceptions,
and timeouts are recorded in diagnostics and do not replace the current
solution.

\paragraph{TTP validation.}
For TTP, generated operators are executed through a hard-timeout subprocess or a
persistent worker process. A route proposal must be a list of length \(n\) that
forms a valid permutation of all cities. A packing proposal must be a binary
list of length \(m\), and the total selected item weight must not exceed the
knapsack capacity. Operators that time out, raise exceptions, return malformed
objects, or violate these feasibility conditions are rejected.

\paragraph{Generation retry.}
When a generated operator fails syntax, registration, timeout, or feasibility
checks, the framework can regenerate the operator by appending the observed
error message to the prompt. Only validated operators are inserted into the
population or used in final testing.

\section{Best Evolved Operator Pairs}
\label{app:best_operator_pairs}

This section reports the evolutionary trajectories and the best evolved
operator pairs selected according to validation performance. The trajectory
figures are included to illustrate how the operator populations evolve across
generations and how representative route/tour-side and packing/purchasing-side
mechanisms emerge during the search process. The final operators shown below
passed syntax validation, interface validation, bounded execution validation,
component-feasibility validation, and complete-solution validation before being
used in final testing. They are included to improve transparency and
reproducibility rather than as hand-designed heuristics. The operators are not
manually tuned after selection; they are the validated outputs selected by the
CoEvo-AHD evolution process.

\subsection{Evolutionary Trajectories and Operator Milestones}
\label{app:evolutionary_trajectories}

\begin{figure}[p]
    \centering
    \vspace*{\fill}
    \includegraphics[width=\textwidth,height=0.76\textheight,keepaspectratio]{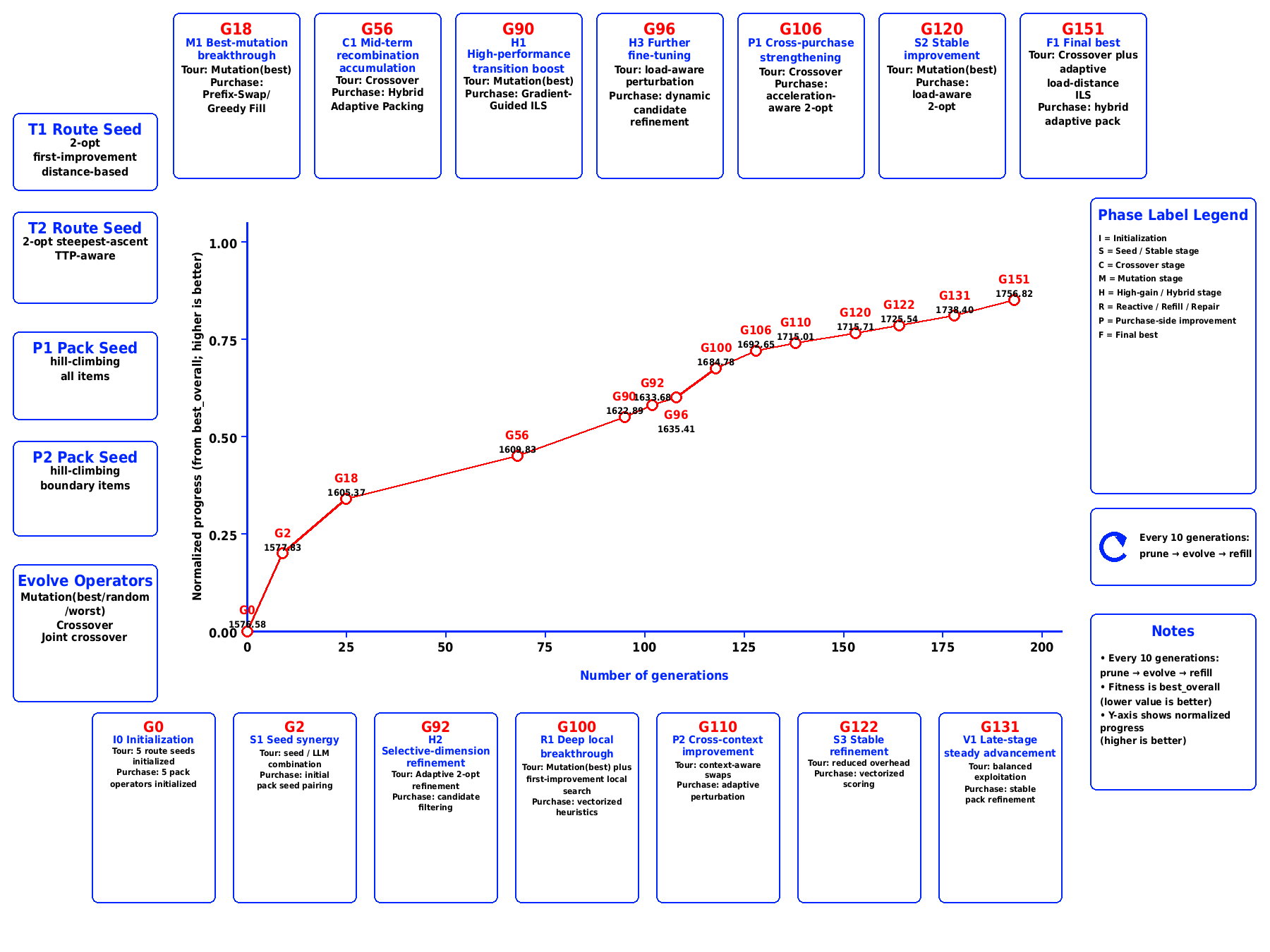}
    \caption{Evolutionary trajectory and representative operator milestones of CoEvo-AHD on TTP. Higher validation performance is better.}
    \label{fig:ttp_evolutionary_trajectory}
    \vspace*{\fill}
\end{figure}

\begin{figure}[p]
    \centering
    \vspace*{\fill}
    \includegraphics[width=\textwidth,height=0.76\textheight,keepaspectratio]{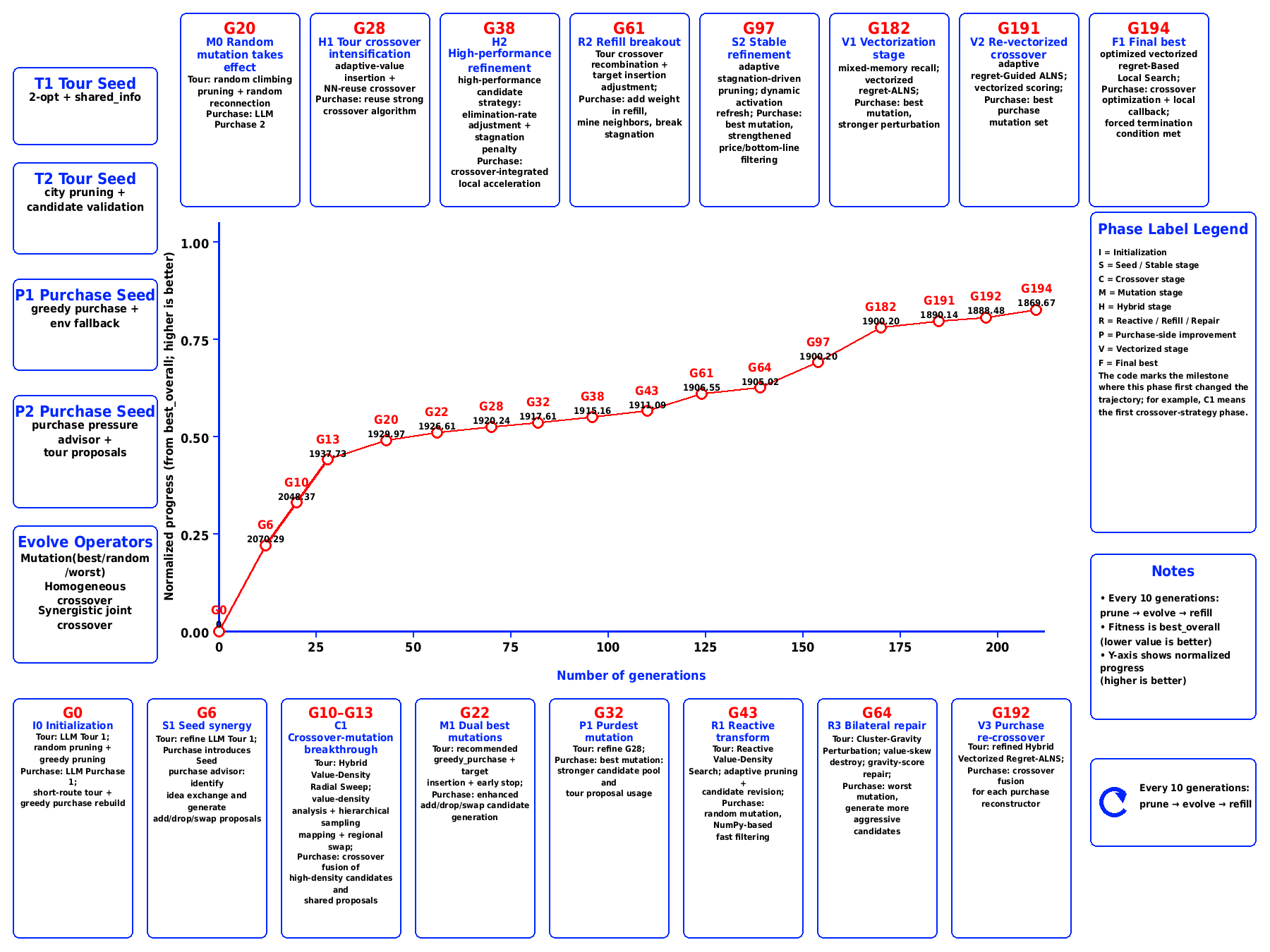}
    \caption{Evolutionary trajectory and representative operator milestones of CoEvo-AHD on TPP. Lower validation cost is better, and the plotted value is normalized as progress toward the final best result.}
    \label{fig:tpp_evolutionary_trajectory}
    \vspace*{\fill}
\end{figure}

Figures~\ref{fig:ttp_evolutionary_trajectory} and
\ref{fig:tpp_evolutionary_trajectory} visualize the evolutionary trajectories
of CoEvo-AHD on TTP and TPP, respectively. The horizontal axis denotes the
number of generations, and the vertical axis denotes the normalized progress
from the initial best validation result to the final best validation result.
The red curve records the best-so-far validation performance during evolution,
whereas the annotated boxes summarize representative operator milestones. To
avoid interrupting the operator listings, these two dense figures are placed on
dedicated appendix pages before the final evolved operator code.

For TTP, the trajectory shows that early improvements are mainly driven by route
and packing seed operators, whereas later improvements are associated with
load-aware route perturbation, route--packing coordination, and adaptive packing
refinement. The final best pair is obtained at generation~151, which is
consistent with the selected TTP operator pair reported in
Appendix~\ref{app:best_ttp_pair}.

For TPP, the trajectory indicates a staged improvement process. Initial gains
come from seed synergy, greedy purchase reconstruction, and shared add/drop/swap
proposals, whereas later gains are associated with crossover-based tour
refinement, purchase-side candidate generation, and vectorized regret-based
search. The final best pair is obtained at generation~194, which corresponds to
the selected TPP operator pair reported in Appendix~\ref{app:best_tpp_pair}.

\clearpage
\clearpage

\clearpage

\subsection{Best Evolved Operator Pair for TTP}
\label{app:best_ttp_pair}

The best evolved TTP operator pair was selected at generation 151. Its validation
average objective is 1756.82, where higher is better, and its
pair-level collaboration score is 1107.21. 
\begin{operatorbox}{Best Evolved TTP Route Operator}
\small
\begin{lstlisting}[style=coevoPython]
import random
import numpy as np

def route_operator(route, pack_plan, problem_data):
    env = problem_data['env']
    items = problem_data['items']
    n_cities = problem_data['n_cities']
    n_items = problem_data['n_items']
    capacity = problem_data['capacity']
    shared = problem_data.get('shared_info', {})

    city_weights = np.zeros(n_cities)
    for k in range(n_items):
        if pack_plan[k]:
            city_weights[items[k][3]] += items[k][2]

    def best_2opt_move(r):
        n = len(r)
        for _ in range(n):
            i = random.randint(0, max(0, n - 2))
            for j in random.sample(range(i + 2, n), max(0, n - i - 2)):
                delta = env.fast_2opt_delta(r, pack_plan, i, j)
                if delta > 1e-9:
                    return i, j
        return None

    def local_search(r):
        r = list(r)
        while True:
            move = best_2opt_move(r)
            if move is None:
                break
            i, j = move
            r[i + 1:j + 1] = reversed(r[i + 1:j + 1])
        return r

    def load_aware_perturb(r, strength):
        r = list(r)
        limit = int(len(r) * 0.65)
        candidates = sorted(
            [(c, city_weights[c] * (idx + 1)) for idx, c in enumerate(r[:limit]) if city_weights[c] > 0],
            key=lambda x: x[1], reverse=True
        )
        for city, _ in candidates[:strength]:
            r.remove(city)
            r.insert(random.randint(limit, len(r)), city)
        return r

    current = local_search(route)
    best_route = current[:]
    best_score = env.evaluate(best_route, pack_plan)
    stagnation = 0

    for _ in range(25):
        candidate = load_aware_perturb(current, min(max(3, n_cities // 12) + stagnation, n_cities // 4))
        candidate = local_search(candidate)
        score = env.evaluate(candidate, pack_plan)
        if score > best_score + 1e-9:
            best_route, best_score = candidate[:], score
            current, stagnation = candidate[:], 0
        else:
            stagnation += 1
            if stagnation >= 8:
                random.shuffle(current)
                current = local_search(current)
                stagnation = 0

    shared['route_position'] = {city: idx for idx, city in enumerate(best_route)}

    load_profile, load = [], 0.0
    for city in best_route:
        load += city_weights[city]
        load_profile.append(load)
    shared['city_load_profile'] = load_profile
    shared['safe_zone_threshold_index'] = int(n_cities * 0.7)
    shared['acceleration_segments'] = [
        (i, i) for i, w in enumerate(load_profile) if w <= 0.3 * capacity
    ]

    return best_route
\end{lstlisting}
\end{operatorbox}

\begin{operatorbox}{Best Evolved TTP Pack Operator}
\small
\begin{lstlisting}[style=coevoPython]
import random
import numpy as np

def pack_operator(route, pack_plan, problem_data):
    env = problem_data['env']
    items = np.array(problem_data['items'], dtype=float)
    capacity = problem_data['capacity']
    shared = problem_data.get('shared_info', {})

    profits, weights, cities = items[:, 1], items[:, 2], items[:, 3].astype(int)
    n_items = len(items)
    route_pos = shared.get('route_position', {c: i for i, c in enumerate(route)})
    positions = np.array([route_pos.get(c, len(route)) for c in cities])

    def score(indices, current_weight, add=True):
        eff = profits[indices] / np.maximum(weights[indices], 1e-9)
        pos_factor = 0.3 + 0.7 * positions[indices] / max(1, len(route) - 1)
        load_penalty = (current_weight + weights[indices]) / max(1.0, capacity)
        return eff * pos_factor - (0.15 * load_penalty if add else -0.15 * load_penalty)

    current = list(pack_plan)
    best_plan = current[:]
    best_obj = env.evaluate(route, best_plan)
    stagnation = 0

    for _ in range(200):
        picked = np.flatnonzero(np.array(current, dtype=bool))
        unpicked = np.flatnonzero(~np.array(current, dtype=bool))
        current_weight = float(np.sum(weights[picked]))
        improved = False

        remove_order = picked[np.argsort(-score(picked, current_weight, add=False))][:15]
        add_order = unpicked[np.argsort(-score(unpicked, current_weight, add=True))][:40]

        for out_idx in remove_order:
            for in_idx in add_order:
                if current_weight - weights[out_idx] + weights[in_idx] <= capacity:
                    trial = current[:]
                    trial[out_idx], trial[in_idx] = 0, 1
                    obj = env.evaluate(route, trial)
                    if obj > best_obj + 1e-7:
                        current, best_plan, best_obj = trial[:], trial[:], obj
                        stagnation, improved = 0, True
                        break
            if improved:
                break

        if not improved:
            stagnation += 1
            if stagnation >= 5 and len(picked) > 0:
                drop_num = min(5, max(1, len(picked) // 4))
                for idx in random.sample(list(picked), drop_num):
                    current[idx] = 0
                stagnation = 0

        if stagnation >= 30:
            break

    return best_plan
\end{lstlisting}
\end{operatorbox}

\subsection{Best Evolved Operator Pair for TPP}
\label{app:best_tpp_pair}

The best evolved TPP operator pair was selected at generation 194. Its validation
average objective is 1869.67, where lower is better, and its
pair-level collaboration score is 1.76.

\begin{operatorbox}{Best Evolved TPP Tour Operator}
\small
\begin{lstlisting}[style=coevoPython]
import random
import numpy as np

def tour_operator(tour, purchasing_plan, problem_data):
    env = problem_data.get('env')
    if env is None:
        return list(dict.fromkeys(tour)) or [0]

    dist = np.array(problem_data['dist_matrix'], dtype=float)
    prices = np.array(problem_data['prices'], dtype=float)
    quantities = np.array(problem_data['quantities'], dtype=float)
    demands = np.array(problem_data['demands'], dtype=float)
    n_cities = problem_data['n_cities']
    shared = problem_data.get('shared_info', {})

    def clean(t):
        t = list(dict.fromkeys(int(c) for c in t)) or [0]
        return [0] + [c for c in t if c != 0]

    def evaluate(t):
        plan = env.greedy_purchase(t)
        return env.evaluate(t, plan), plan

    def best_insertion(current_tour):
        tour_set = set(current_tour)
        t = np.array(current_tour, dtype=int)
        if len(t) < 2:
            return None

        tour_mask = np.zeros(n_cities, dtype=bool)
        tour_mask[t] = True
        tour_mask[0] = False

        curr_prices = prices[:, tour_mask].copy()
        curr_qty = quantities[:, tour_mask]
        curr_prices[curr_qty <= 0] = np.inf
        current_best = np.min(curr_prices, axis=1)

        candidates = np.array([c for c in range(1, n_cities) if c not in tour_set], dtype=int)
        if len(candidates) == 0:
            return None

        cand_prices = prices[:, candidates].copy()
        cand_prices[quantities[:, candidates] <= 0] = np.inf
        savings = np.maximum(current_best[:, None] - cand_prices, 0.0) * demands[:, None]
        total_savings = np.sum(savings, axis=0)

        u, v = t[:-1], t[1:]
        delta = dist[candidates[:, None], u] + dist[candidates[:, None], v] - dist[u, v]
        best_pos = np.argmin(delta, axis=1) + 1
        net_change = np.min(delta, axis=1) - total_savings

        idx = int(np.argmin(net_change))
        if net_change[idx] >= 0:
            return None
        return int(candidates[idx]), int(best_pos[idx])

    current = clean(tour)
    best_cost, best_plan = evaluate(current)
    best_tour = current[:]
    no_improve = 0

    for _ in range(50):
        move = best_insertion(current)
        if move is not None:
            city, pos = move
            candidate = current[:pos] + [city] + current[pos:]
            cost, plan = evaluate(candidate)
            if cost < best_cost - 1e-9:
                current, best_tour, best_cost, best_plan = candidate[:], candidate[:], cost, plan
                no_improve = 0
                continue

        no_improve += 1
        if no_improve >= 10:
            current = best_tour[:]
            removable = [i for i, c in enumerate(current) if c != 0]
            for idx in sorted(random.sample(removable, max(1, len(removable) // 3)), reverse=True):
                del current[idx]
            no_improve = 0

    shared['useful_cities'] = [c for c in best_tour if c != 0]
    shared['tour_strategy'] = 'regret_insertion_ils_core'
    return best_tour
\end{lstlisting}
\end{operatorbox}

\begin{operatorbox}{Best Evolved TPP Purchase Operator}
\small
\begin{lstlisting}[style=coevoPython]
import numpy as np

def purchase_operator(tour, purchasing_plan, problem_data):
    prices = np.array(problem_data['prices'], dtype=float)
    quantities = np.array(problem_data['quantities'], dtype=float)
    demands = np.array(problem_data['demands'], dtype=float)
    dist = np.array(problem_data['dist_matrix'], dtype=float)
    env = problem_data.get('env')
    shared = problem_data.get('shared_info', {})

    base_tour = [0] + [c for c in dict.fromkeys(tour) if c != 0]
    tour_set = set(base_tour)
    tour_arr = np.array(base_tour, dtype=int)

    base_plan = env.greedy_purchase(base_tour) if env else {}

    used = {c for buys in base_plan.values() for c, q in buys if q > 0}
    idle_cities = [c for c in base_tour if c != 0 and c not in used]

    def insertion_cost(city):
        if city in tour_set or len(tour_arr) < 2:
            return 0.0
        u, v = tour_arr[:-1], tour_arr[1:]
        return float(np.min(dist[u, city] + dist[city, v] - dist[u, v]))

    tour_best = np.full(len(demands), np.inf)
    for p in range(len(demands)):
        available = quantities[p, list(tour_set)] > 0
        if np.any(available):
            tour_best[p] = np.min(prices[p, list(tour_set)][available])

    regret_scores = {}
    for c in range(prices.shape[1]):
        if c in tour_set:
            continue
        mask = quantities[:, c] > 0
        savings = np.maximum(tour_best[mask] - prices[mask, c], 0.0) * demands[mask]
        total_saving = float(np.sum(savings))
        if total_saving > 0:
            regret_scores[c] = total_saving / max(insertion_cost(c), 1e-9)

    add_cities = [c for c, _ in sorted(regret_scores.items(), key=lambda x: -x[1])[:10]]
    drop_cities = idle_cities[:10]
    swap_moves = [(d, a) for d in drop_cities[:5] for a in add_cities[:5] if a not in tour_set][:15]

    shared['add_cities'] = add_cities
    shared['drop_cities'] = drop_cities
    shared['swap_moves'] = swap_moves

    return base_plan
\end{lstlisting}
\end{operatorbox}

\section{Experimental Details}
\label{app:experimental_details}

\subsection{Baseline Settings}
\label{app:baseline_settings}

We compare CoEvo-AHD with representative baseline algorithms for both TTP and
TPP. All methods are evaluated under the same final testing budget whenever
applicable. For TPP, all baselines are adapted to the same instance parser,
purchasing reconstruction procedure, objective evaluator, and output format.
For a fixed set of visited markets, the purchasing plan is recomputed using the
same price-supply-demand information, and the final objective value is evaluated
by the unified TPP evaluator. Therefore, the comparison reflects differences in
search strategies rather than differences in objective implementation.

\subsubsection{TPP Baselines}

For TPP, we compare CoEvo-AHD with three representative metaheuristic baselines:
DMD-ATA, ALNS, and Transgenetic Algorithm. In the main result table, DMD-ATA is
reported as DMD-ATA (ACO), and Transgenetic Algorithm is reported as TA.

\begin{table}[H]
\centering
\caption{TPP baseline methods and their main settings.}
\label{tab:tpp_baseline_settings}
\small
\begin{tabularx}{\textwidth}{
>{\raggedright\arraybackslash}p{3.3cm}
>{\raggedright\arraybackslash}X
>{\raggedright\arraybackslash}p{4.8cm}}
\toprule
Method & Description & Main parameters \\
\midrule
DMD-ATA (ACO) &
Ant-colony-based TPP solver with multiple pheromone levels, local search, and Dropstar-based intensification. &
$n_{\mathrm{levels}}=30$, evaporation $=0.001$, local search rounds $=2$, Dropstar exact mode, beam size $=24$, time limit $=100$s \\
\midrule
ALNS &
Adaptive large neighborhood search with destroy-repair operators, simulated annealing acceptance, and adaptive operator scoring. &
$T_0=5000$, cooling rate $=0.95$, segment length $=100$, reaction factor $=0.40$, $\beta\in[0.10,0.40]$, time limit $=100$s \\
\midrule
Transgenetic Algorithm (TA) &
Population-based search with host repository, plasmid operations, and transposon operations. &
population size $=50$, $k_{\mathrm{info}}=30$, stage length $b=4$, time limit $=100$s, penalty weight $=5000$ \\
\bottomrule
\end{tabularx}
\end{table}

\paragraph{DMD-ATA (ACO).}
DMD-ATA is implemented as an ant-colony-based TPP solver with multiple pheromone
levels, heterogeneous ant parameters, local search, and Dropstar-based
intensification. Each solution is represented as a market visiting sequence
starting from depot 0. During construction, the next market is selected by
jointly considering pheromone intensity, travel distance, and the potential
improvement in unsatisfied demand or purchasing cost. After a complete solution
is constructed, local search is applied, including 2-opt, market insertion,
simplification, and market drop. Since the original DMD-ATA was mainly designed
for uncapacitated TPP, our capacitated implementation uses Dropstar as a
candidate generation and intensification mechanism. All candidate solutions are
finally evaluated by the unified capacitated TPP evaluator.

\paragraph{ALNS.}
The ALNS baseline follows an adaptive large neighborhood search framework with
destroy and repair operators, simulated annealing acceptance, and adaptive
operator scoring. A solution state consists of a visited-market path and a
purchasing plan. For a fixed market set, the purchasing subproblem is solved by
purchasing each product from visited markets in ascending unit-price order until
the demand is satisfied or available supply is exhausted. The implementation
uses random market removal, random product-market removal, worst purchase-cost
removal, price-similarity removal, distance-gap removal, and high-quantity
removal as destroy operators. The repair operators include greedy product
insertion, randomized product insertion, regret product insertion, and cheapest
market insertion. Since our experiments consider the standard capacitated TPP,
release-time, perishability, deterioration-cost, and cold-chain terms from
TPP-PF variants are removed, leaving only travel cost and purchasing cost.

\paragraph{Transgenetic Algorithm (TA).}
The Transgenetic Algorithm baseline follows a host-repository, plasmid, and
transposon search mechanism. A chromosome represents a market sequence without
the depot, and it is converted into a depot-starting route during evaluation.
The initial population is generated by random feasible construction followed by
unused-market removal and 2-opt route improvement. The host repository stores
both prior path information and high-quality chromosomes from the current
population. Plasmid operations inject promising information strings into
chromosomes, while transposon operations perturb a chromosome by moving or
removing route segments. The original Lin-Kernighan local improvement step is
replaced by deterministic first-improvement 2-opt to avoid external LK/LKH
dependencies. This implementation difference is explicitly reported for
reproducibility.

\subsubsection{TTP Baselines}

For TTP, we compare CoEvo-AHD with MATLS, S5, and CoCo. The detailed parameter
settings of these baselines will be reported after the final TTP baseline
configuration is fixed.

\begin{table}[H]
\centering
\caption{TTP baseline methods and their main settings.}
\label{tab:ttp_baseline_settings}
\small
\begin{tabularx}{\textwidth}{
>{\raggedright\arraybackslash}p{3.3cm}
>{\raggedright\arraybackslash}X
>{\raggedright\arraybackslash}p{4.8cm}}
\toprule
Method & Description & Main parameters \\
\midrule
MATLS &Memetic algorithm for TTP with population-based recombination, TSP-oriented tour improvement, and knapsack-oriented packing local search. Initial tours are generated by CLKSolver when available or by MST-based construction. &
population size $=30$, $500$ MA iterations\\
\midrule
S5 & Restart-based heuristic for TTP that repeatedly generates a TSP tour and applies an iterative greedy packing procedure with parameter search. &
power start $=5.0$, power spread $=2.5$, maximum power-search iterations $=20$, convergence tolerance $=0.1$, $500$ outer iterations\\
\midrule
CoCo & Collaborative local-search heuristic for TTP that alternates between tour improvement and packing-plan improvement. Initial packing is selected from insertion and iterative greedy packing heuristics. &
PackIterative packing with $c=5.0$, $d=2.5$, $q=20$, default algorithm setting $\mathrm{alg}=4$, $500$ inner iterations and \\
\bottomrule
\end{tabularx}
\end{table}
\paragraph{MATLS}
MATLS is a memetic algorithm for the Traveling Thief Problem,
where each individual jointly encodes a city tour and an item-picking plan. The
initial population is generated using CLKSolver-based TSP tours when available,
with an MST-based construction over Delaunay-neighbor edges as fallback, and
random feasible packing plans. During evolution, two parents are selected to
produce an offspring by order crossover for the tour and single-point crossover
for the packing plan, followed by a repair step if the knapsack capacity is
violated. Each offspring is further improved by a two-stage local search: a
2-opt procedure for tour improvement and a single-flip packing search based on
estimated item gains under the current tour. Candidate offspring are inserted
only when sufficiently different from existing individuals, and the population
is sorted by the TTP objective, i.e., collected profit minus renting cost, before
being truncated to the fixed population size. In our implementation, the main
settings are population size $=30$, maximum initialization trials $=100$,
no-improvement limit $=2000$, and, for TTP
instances, $500$ MA iterations with $3$ independent runs per instance.

\paragraph{S5}
S5 is a restart-based heuristic for the Traveling Thief Problem.
In each outer iteration, the method first generates a closed TSP tour using
CLKSolver and then constructs a packing plan with an iterative greedy packing
procedure. For a fixed tour, items are ranked by a distance-aware profit-weight
score, where the heuristic value is proportional to
$\mathrm{profit}^{\alpha}/(d_{\mathrm{to\_go}}\cdot \mathrm{weight}^{\alpha})$.
The exponent $\alpha$ is tuned by a ternary-search-like procedure initialized
at $\alpha=5.0$ with search spread $2.5$, using at most $20$ search iterations
and a convergence tolerance of $0.1$. For each candidate value of $\alpha$,
items are greedily inserted into the packing plan while respecting the knapsack
capacity, and the TTP objective is evaluated as the collected profit minus the
renting cost induced by the load-dependent travel time. The best solution over
all generated tours and packing trials is retained. In our implementation, S5
can be terminated either by a runtime budget or by an outer-iteration limit.

\paragraph{CoCo}
CoCo is  a collaborative local-search heuristic for the Traveling
Thief Problem, where each solution consists of a city tour and a binary packing
plan. The initial tour is generated by CLKSolver when available, and the initial
packing plan is selected as the better one between an insertion-based packing
heuristic and an iterative greedy packing heuristic. The iterative packing
procedure ranks items using a ratio- and distance-aware score and searches the
weighting parameter with initial value $c=5.0$, spread $d=2.5$, and at most
$q=20$ search rounds. Starting from the initial solution, CoCo repeatedly
alternates between tour and packing optimization. The tour is improved by a
steepest-ascent 2-opt search restricted to Delaunay-neighbor candidate moves,
while the packing plan is improved by an all-item bit-flip hill-climbing
procedure that accepts flips only when they increase the TTP objective. Each
solution is evaluated by the collected profit minus the renting cost under the
load-dependent travel speed. The best solution over all restarts and alternating
local-search iterations is retained. In our implementation, the default algorithm
setting is $\mathrm{alg}=4$.
\paragraph{Fairness and implementation notes.}
All baseline methods are evaluated on the same test instances as CoEvo-AHD.
Whenever applicable, the same wall-clock time limit is used. For stochastic
methods, independent repeated runs are conducted and the mean objective value is
reported. For TPP baselines, all final solutions are re-evaluated by the unified
TPP evaluator. Infeasibility penalties are used only during search or repair;
reported final objective values are computed from the corresponding validated
solutions.



\end{document}